\documentclass{article}

\usepackage[preprint]{neurips_2026}

\usepackage{microtype}
\usepackage{graphicx}
\usepackage{tcolorbox}
\tcbuselibrary{listings,breakable}
\usepackage{subcaption}
\usepackage{booktabs} % for professional tables
\usepackage{multirow}
\usepackage[utf8]{inputenc} % allow utf-8 input
\usepackage[T1]{fontenc}    % use 8-bit T1 fonts
\usepackage{hyperref}       % hyperlinks
\usepackage{url}            % simple URL typesetting
\usepackage{booktabs}       % professional-quality tables
\usepackage{amsfonts}       % blackboard math symbols
\usepackage{nicefrac}       % compact symbols for 1/2, etc.
\usepackage{microtype}      % microtypography
\usepackage{xcolor}         % colors
\usepackage{amsmath}
\usepackage{amssymb}
\usepackage{mathtools}
\usepackage{amsthm}
\usepackage{xcolor,colortbl}
\usepackage[capitalize,noabbrev]{cleveref}
\newcommand{\ours}{\texttt{ExpG}}
\definecolor{gainbg}{RGB}{230,245,230}   % lighter green background
\definecolor{darkgreen}{RGB}{0,110,60}     % dark green text
% ...
\definecolor{lightblue}{RGB}{102, 167, 208}
\definecolor{darkred}{RGB}{139,0,0}
\makeatletter
\renewcommand\@fnsymbol[1]{\ensuremath{\ifcase#1\or *\or \dagger\or \ddagger\or \mathsection\or \mathparagraph\or \|\or **\or \dagger\dagger\or \ddagger\ddagger \else\@ctrerr\fi}}
\makeatother
\theoremstyle{plain}
\usepackage{url}

\theoremstyle{definition}

\theoremstyle{remark}

\usepackage[ruled,vlined]{algorithm2e}

\title{Towards Robust Tool Use in Agents \\ via Experience-Driven Adaptive Guidance}

\author{%
\textbf{Can Wang\textsuperscript{\rm 1,2},
Haoran Chen\textsuperscript{\rm 2},
Li Yu\textsuperscript{\rm 2},
Ding Hao\textsuperscript{\rm 3},
Bohai Zhao\textsuperscript{\rm 1},
Zhaoyang Liu\textsuperscript{\rm 2},
Zhiying Tu\textsuperscript{\rm 1}}
\\[0.12in]
\textsuperscript{\rm 1}Shandong Key Laboratory of Digital Service Computing Technology and Systems \\
\textsuperscript{\rm 2}Alibaba Token Hub \\
\textsuperscript{\rm 3}Department of Electrical and Electronic Engineering, The Hong Kong Polytechnic University
\vspace{0.18in}
}

\begin{document}

\maketitle

\begin{abstract}
The performance bottleneck of agents is increasingly shifting from model capability to the robustness of their execution processes. 
Tools play a central role as the primary interface through which agents interact with external environments, yet existing methods rarely focus on ensuring robust tool use across diverse runtime conditions. 
To address this problem, we propose {\ours}, a mechanism that builds and refines adaptive guidance capturing each tool’s capability boundaries and best practices, thereby enabling agents to use tools more robustly and effectively. 
{\ours} consists of three phases: 
(1) experience acquisition, which analyzes tool invocation quality from historical execution trajectories, producing structured learnable experiences through multi-aspect attribution; 
(2) experience distillation, which keeps the experience pool effective by filtering unhelpful experiences, selecting representative ones with an equivalence-class-based method, and summarizing them into generalizable guidance; and 
(3) experience reuse, which applies the guidance adaptively during future task solving. 
Extensive experiments show that {\ours} brings consistent improvements across the tool selection, tool calling, and response generation tasks, enabling smaller agents to outperform larger ones that do not use {\ours}. Moreover, {\ours} achieves particularly strong gains in challenging settings, suggesting a promising path toward more robust tool use. 
Our code, experiments, and results are available at \href{https://github.com/WangCan1178/ExpG}{\texttt{https://github.com/WangCan1178/ExpG}}. 
\end{abstract}

\section{Introduction}

% By invoking external tools, agents can tap into domain-specific expertise, access continuously updated real-time data and interact with users in more flexible ways~\cite{1,3,5}. 
% Equipped with a rich ecosystem of external tools, agents overcome the limitations of their inherent knowledge and reasoning capabilities, thereby handling tasks with greater accuracy, reliability, and efficiency.
% Tool empowerment represents an important milestone in the evolution of large language models (LLMs) toward truly intelligent agents. Equipped with a rich ecosystem of external tools, agents can tap into domain-specific expertise, access continuously updated real-time data, and interact with users in more flexible ways~\cite{1,3,5}, thereby overcoming the limitations of their inherent knowledge and reasoning capabilities to handle tasks more effectively. 
% 智能体整体性能的瓶颈正在从模型本身转移到其依赖的执行基础设施，也就是 harness。Harness 通过运行时控制提供行为治理与容错能力，使智能体能够在长时任务中保持稳定并在动态环境中自我修正。工具调用是 harness 最核心的治理点之一。已有研究显示，在模型保持不变的情况下，仅调整 harness 中的工具接口格式，就能在 code patch 任务中带来六到十倍的性能差异，由此体现出 harness 对系统表现的决定性影响

% The performance bottleneck of agents is shifting from the large language model (LLM) to the harness layer~\cite{1h,2h,4h}. The harness provides runtime control that enables behavioral governance and fault tolerance, allowing agents to remain stable over complex tasks and adapt to dynamic environments. Research~\cite{3h} shows that tool use is a core focus of the harness: even without changing the agent itself, simply modifying the tool interface can lead to six to ten times better performance in code patching. 
The performance bottleneck of agents is shifting from the capabilities of large language models (LLMs) to the robustness of their execution processes. Agent harnesses~\cite{1h,2h,4h} have therefore been proposed to provide runtime control that enables behavioral governance and fault tolerance, allowing agents to remain stable when performing complex tasks in dynamic environments.   
Within such harnesses, tools serve as the primary interface through which agents interact with external environments~\cite{1,5}, and prior studies~\cite{3h} show that even minor improvements in tool-use robustness can yield performance gains ranging from 6.7\% to 68.3\%.
% , highlighting the importance of reliable tool use for agent execution.

% Existing tool-use methods still focus on static environments and rely on an implicit assumption that tool-use behavior is predictable. 
% On the one hand, these work still focus on static environments and rely on an implicit assumption that tool-use behavior is predictable. this assumption implies that tools the agent has already mastered can always be invoked correctly and return expected responses. In reality, the quality of tool invocations is strongly affected by environmental dynamics~\citep{7,9}. Even for the same tool, the agent’s behavior and answer quality can change with the task context, and may also drift over time or across different usage environments. This issue becomes particularly evident when there is a gap between development and production: many tools that work well in the original setting become unreliable or even unusable because their previous usage patterns no longer apply~\citep{10}.
% On the other hand, the assumption also implies that even when tools are invoked incorrectly, their failures are accompanied by informative signals, enabling the agent to continuously self-reflect and eventually learn to use the tools correctly. In fact, the responses returned by tool invocations are often noisy and coarse-grained~\cite{8}. For instance, a buggy calculator may return an incorrect numerical result while still reporting the invocation as “successful”. Figure~\ref{fig1}(a) presents examples of these two challenges.

\begin{figure*}[tbp]
    \centering
    \includegraphics[width=0.98\linewidth]{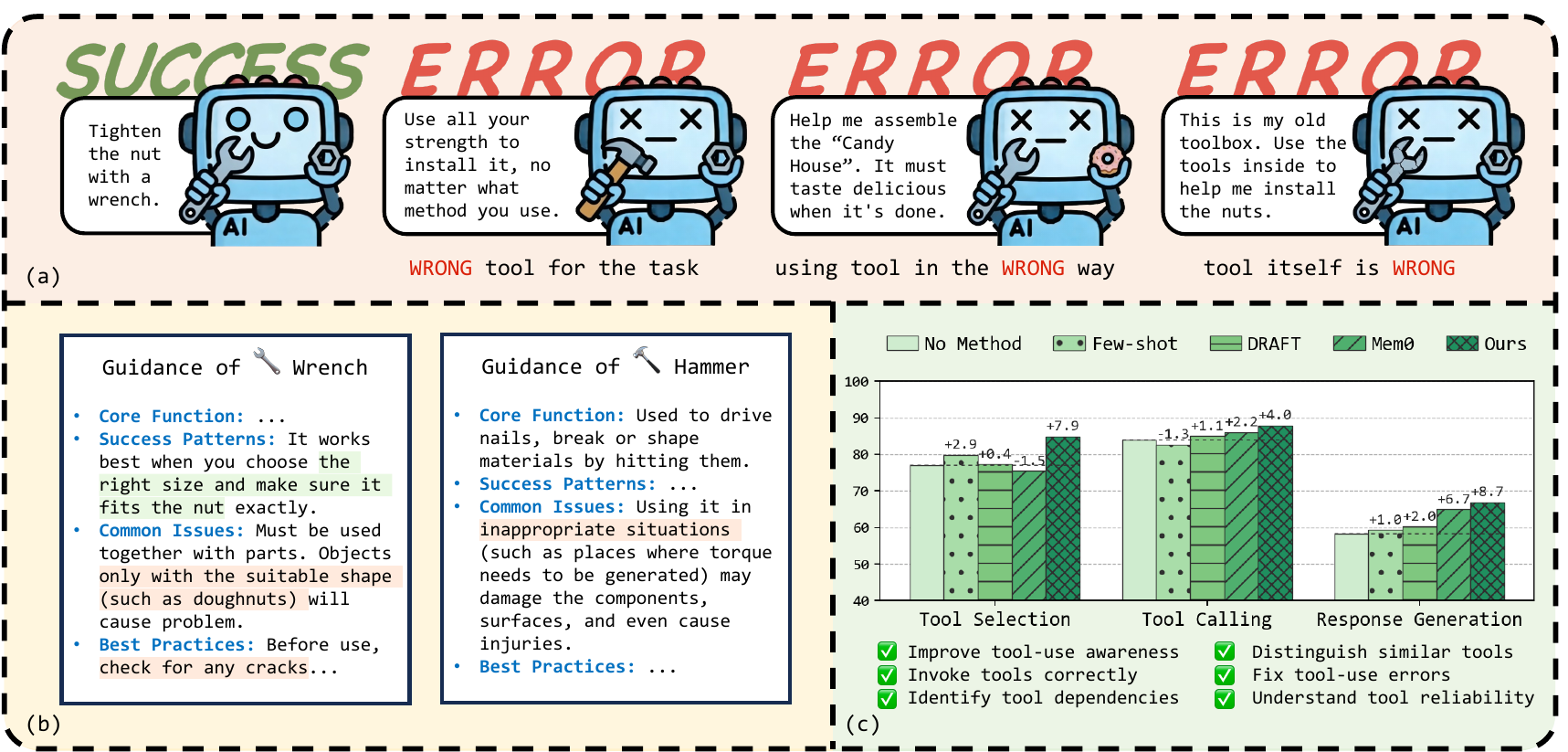}
    \caption{(a) Two real-world challenges: the agent uses the same tool but gets different results due to the different context; there are different reasons for the error, but the tools respond with the same coarse-grained signal (both return just ``error''). (b) Guidance learned from past tool‑use experience. (c) Results across different stages of tool usage demonstrate that {\ours} improves agent tool-use performance from multiple aspects.}\label{fig1} 
    % \vspace{-4mm}
\end{figure*}

However, existing tool-use methods~\cite{toolsurvey} rarely consider robustness across diverse runtime environments, as they typically developed under conditions where tool-use behavior remains predictable.
% as they are primarily designed for static situations where tool-use behavior is implicitly assumed to be predictable. 
Concretely, it implies that (1) tools the agent has already mastered can always be invoked correctly and return expected responses, and (2) even when tools are invoked incorrectly, their failures are accompanied by informative signals, enabling the agent to continuously self-reflect and eventually learn to use the tools correctly. 
Unfortunately, real-world conditions break these assumptions. On the one hand, the quality of tool invocations is strongly affected by environmental dynamics~\citep{7, 9}. Even for the same tool, the agent’s tool-use behavior can change with the task context, and may also drift over time or across different usage environments. This is particularly evident when there is a gap between development and production: many tools that work well in the original setting become unreliable or even unusable because their previous usage patterns no longer apply~\citep{10}. 
On the other hand, the responses from tool invocations are often noisy or coarse-grained~\cite{8}. For instance, a buggy calculator may return an incorrect numerical result while still reporting the invocation as ``successful''. Figure~\ref{fig1}(a) illustrates examples of these two challenges. 
Without a clear understanding of a tool’s capability boundaries and how to use it robustly under varying conditions, agents struggle to generalize tool use reliably across diverse environments and often fall back on trial and error. 
Once a mistake occurs, the agent may receive only ambiguous signals, which can lead to an increasingly complicated and often incorrect inference path.

% To address these challenges, we propose {\ours} (Dynamic Experience-Driven experience pool), a mechanism that dynamically builds capability boundaries and best practices for tools when agents invoke them blindly. 
% Inspired by agent memory~\cite{memsurvey}, {\ours} acquires, distills, and reuses past experience to produce a unique ``memory'' for each tool, which serves as prior guidance to help the agent interact with the tool more effectively.
% (experiences pool that can be dynamically updated)
To address these challenges, we propose {\ours}, a mechanism that produces \textbf{Exp}erience-Driven adaptive \textbf{G}uidance 
% to help agents use tools effectively under environmental dynamics and noisy feedback. 
to help agents use tools effectively in the presence of environmental dynamics and noisy feedback.
% across diverse environments. 
% ensure its. unpredictable
% which serves as prior guidance to help the agent interact with the tool more effectively.
% Specifically, in contrast to existing approaches that rely on static assumptions and treat tool invocations as black-box, isolated, and perfectly
In contrast to existing approaches that rely on static assumptions and treat tool invocations as black-box, isolated, and repeatable events, our work views tool invocations as analyzable, interdependent, and learnable experiences. 
% We partition experiences into equivalence classes aligned with invocation patterns so that guidance remains representative yet covers rare failure modes across acquisition, distillation, and reuse. 
% from past tool invocations
% Inspired by agent memory~\cite{memsurvey}, {\ours} maintains an evolving experience pool by continuously acquiring, distilling, and reusing these experiences, continually building and refining the generated guidance to help agents use tools effectively under environmental dynamics and noisy feedback. 
% By continuously acquiring, distilling, and reusing these experiences, {\ours} maintains an evolving experience pool, inspired by agent memory~\cite{memsurvey}, thus adaptively building and refining guidance for tool use. 
By continuously acquiring, distilling, and reusing these experiences, {\ours} maintains an evolving experience pool inspired by agent memory~\cite{memsurvey}, which enables it to adaptively build and refine guidance for tool use.
% generalizable
% The guidance characterizes tools' capability boundaries and best practices, helping agents using tools from 
% tool invocation 
Specifically, in the first phase, {\ours} acquires experiences by analyzing agents’ execution trajectories and attributing each tool invocation’s success or failure to contributing factors. Experiences related to the same tool are categorized into different invocation patterns through equivalence-class partitioning, resulting in an initial structured experience pool. 
% These experiences are partitioned into equivalence classes, such that each class corresponds to a distinct tool invocation pattern. 
A filtering method is then applied to maintain the quality of the experience pool by removing unhelpful experiences and selecting helpful ones, 
% It builds on the partitioning of the acquired experiences into equivalence classes, 
thereby retaining representative experiences covering as many invocation patterns as possible while preserving the original distribution. 
% that capture each tool’s common success and failure patterns. 
% In contrast to existing approaches that rely on static assumptions and treat tool invocations as black-box, isolated, and perfectly repeatable events, our work views tool invocations as analyzable, interdependent, and learnable experiences. {\ours} acquires these experiences in the first phase by attributing the success and failure of tool invocations from agents’ execution trajectories. 
% Our designed selecting method based on equivalence classes is then applied to promote diversity while preserving the original experience distribution, retaining representative experiences that capture each tool’s common success and failure patterns. 
With these filtered experiences, {\ours} performs qualitative behavior analysis alongside quantitative performance analysis to construct guidance for each tool, organizing fragmented invocation-level experiences into generalizable tool-level guidance. 
The guidance characterizes the capability boundaries and best practices of tools, and can be adaptively refined as the experience pool evolves. 
% which can be continuously refined as more experiences are acquired. 
Finally, the guidance is reused in subsequent tasks whenever the agent invokes the same tool, either as dynamic contextual prompts or as stable schema constraints, thereby improving task performance in complex and changing real-world applications through more robust tool use. 
% enabling more robust and reliable tool use and improving task performance in complex and unpredictable real-world environments.
% \zhaoyang{Illustrate more about our method contribution and innovation?}

In summary, our contributions are as follows:
\begin{itemize}%[leftmargin=12pt,topsep=2pt,itemsep=2pt]
    \item  
    % To overcome the challenges posed by reliance on assumption of predictable tool usage,
    % To overcome unpredictable tool-use behaviors caused by environmental dynamics and noisy feedback, 
    % we break out of the static view of tool invocation by characterizing them through a multi‑faceted attribution method, thereby acquiring learnable experience that encapsulates comprehensive information about the environment, the agent, and the tool itself.
    To overcome unpredictable tool-use behaviors caused by environmental dynamics and noisy feedback, we break out of the static view of tool invocation by introducing a structured characterization of tool invocations, thereby acquiring learnable experiences through equivalence‑class partitioning that encapsulate comprehensive information about the environment, the agent, and the tool itself.
    % \item We identify two key challenges posed by existing methods' reliance on the assumption of perfect tool invocation environments. We identify the static assumptions commonly relied upon in existing methods for enhancing agent tool use and highlight the challenges these assumptions face in real-world deployment environments. To overcome these challenges, we shift the perspective from viewing tool invocations as black-box, static, and perfectly repeatable isolated events. Instead, we characterize tool invocations by considering the context, the agent, and the tool itself, enabling the acquisition of insightful experiences from agent-tool interactions to improve subsequent tool use. 
    
    \item We propose a mechanism \textbf{\ours} to maintain an evolving experience pool at the tool level. By acquiring, distilling, and reusing past experiences, {\ours} builds capability boundaries and best practices for tools, which serve as practical guidance and can be continually refined as the agent becomes more effective in tool usage. 
    % We propose a dynamic mechanism \textbf{\ours} to maintain memory at the tool level. By acquiring, distilling, and reusing past experiences, {\ours} builds capability boundaries and best practices for tools, which serve as guidance and can be continually refined as the agent becomes more efficient in tool invocation. 
    % builds capability boundaries and best practices for tools, which serve as guidance
    % As the agent continuously interacts with tools, the guidance is dynamically refined, facilitating mutual improvement between the agent and the tools.
    % in the presence of environmental noise, coarse-grained feedback, and imperfect execution
    \item 
    % We conduct extensive experiments covering different stages of tool use, validating the effectiveness of {\ours} across various models, showing that small models equipped with memory can outperform larger memoryless ones. In particular, {\ours} performs notably well under challenging settings, such as noisy environments provided with distractor tools or imperfect information. An in-depth case analysis further demonstrates that {\ours} improves tool-use performance across multiple aspects.
    Extensive experiments across different stages of tool use validate the effectiveness of our approach across a range of models, showing that small models equipped with {\ours} can outperform larger ones without it. Notably, {\ours} remains robust under challenging settings, such as noisy environments with distractor tools or imperfect information. 
    % An in-depth case analysis further demonstrates that {\ours} improves tool-use performance across multiple aspects.
    % faced with practical challenges. 
    % We conduct extensive experiments in tool-use scenarios characterized by environmental dynamics and noise to systematically evaluate the effectiveness of ReMe. Experimental results demonstrate that ReMe enhances overall performance across the entire process of tool use, significantly improving the agent's robustness and task success rate. 
    % Notably, ReMe exhibits particular advantages in distinguishing between functionally similar or semantically close tools and in correcting unreliable or suboptimal tool invocations, markedly improving the agent's tool-use effectiveness in complex, near-real-world environments.
    % 这种闭环的机制使得智能体能够从历史经验中持续学习，不断优化其工具选择、参数配置和使用策略，最终实现工具使用能力的渐进式提升。
\end{itemize}

\section{Related Works}
\textbf{Tool-augmented LLMs.} 
Tool-augmented LLMs substantially enhance domain expertise~\cite{1,2}, knowledge acquisition~\cite{5,6}, and interaction quality~\cite{3,4}, and are seen as a key step in the evolution from pure language models toward intelligent agents. Existing methods for improving tool use can be divided into two categories based on whether model parameters are explicitly updated~\cite{toolsurvey}.
% : tuning-based and tuning-free approaches. 
Tuning-based methods update model parameters to internalize tool-use policies. They typically rely on supervised learning~\cite{11,12} or reinforcement learning~\cite{14,15}, using pre-collected data and tool-execution feedback to train LLMs. In contrast, tuning-free methods keep model parameters fixed and instead improve tool use by guiding the LLM during inference, for example through carefully designed prompts~\cite{16}, chain-of-thought prompting~\cite{17}, and the ReAct paradigm~\cite{13}. 
Recent studies have highlighted robustness challenges of tool-augmented LLMs in real-world settings~\cite{10,8,7,9}.
Building on this observation, 
we construct guidance that can be continually refined from structured experiences, enabling tool-augmented agents to cope better with unpredictable environments.

\noindent
\textbf{Harness Engineering.} 
As LLM become increasingly capable, research attention has gradually shifted from improving model weights to engineering the surrounding agent infrastructure that enables reliable deployment\cite{18,19}. In particular, harness engineering~\cite{1h,2h} has emerged as a key direction for improving the robustness of LLM agents. The execution harness~\cite{3h,4h} governs how an agent interacts with its environment by managing the execution loop, tool access, context construction, state persistence, and runtime monitoring. 
Harness optimization can be achieved through multiple external components such as memory stores~\cite{memsurvey, chhikara2025mem0}, reusable skills~\cite{1s, 2s}, and context engineering~\cite{c1, c2}, which respectively improve information persistence, procedural reuse, and structured task execution during runtime. Most existing work improves robustness by externalizing capabilities at the task or workflow level within the harness. In contrast, {\ours} operates at the fine-grained granularity of the tool, externalizing the limits per tool capability and best practices that are complementary to methods based on higher level memory or skills.

\begin{figure*}[t]
    \centering
    \includegraphics[width=0.98\linewidth]{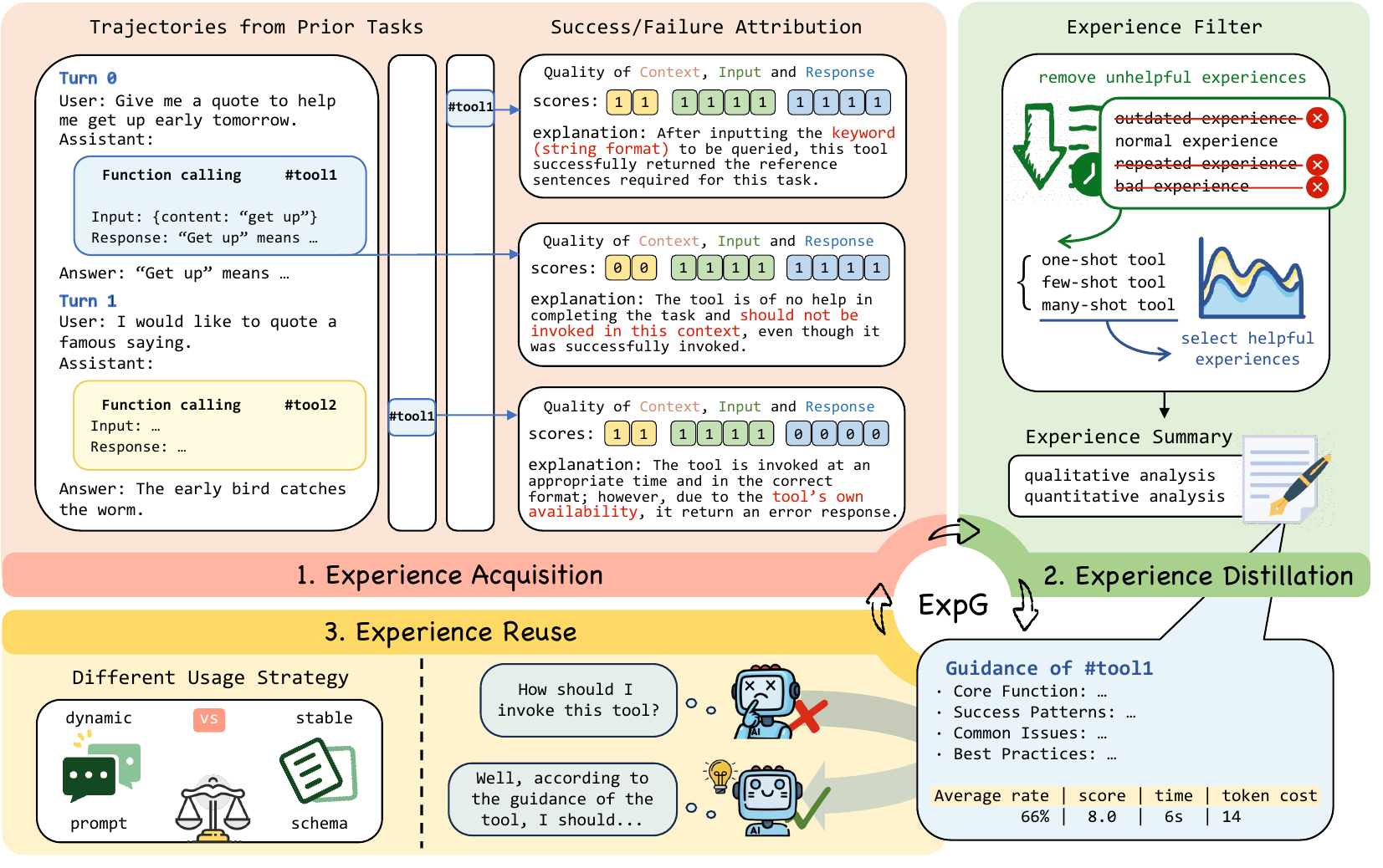}
    \caption{Overview of {\ours}. It consists of three phases: acquiring tool invocation experiences from past trajectories, distilling them into guidance, and reusing them to improve future task performance. These phases form a mechanism that builds capability boundaries and best practices for each tool.}\label{fig:overview}
    \vspace{-5mm}
\end{figure*}

% , which can be continuously updated and refined as if the agent has ``memory'' for using the tool effectively.
\section{Methodology}\label{method}
\subsection{Overview of \ours}
Our work, \ours, as illustrated in Figure~\ref{fig:overview}, operates through three sequential phases: experience acquisition, distillation, and reuse. In the \textbf{experience acquisition} phase, an evaluator attributes the success or failure of each tool invocation from historical trajectories, acquiring structured and learnable experiences. In the \textbf{experience distillation} phase, experiences are first filtered to ensure they are helpful, then summarized into guidance that captures the tool's capability boundaries and best practices. Finally, the \textbf{experience reuse} phase injects these distilled guidance into the agent’s inference process, either as part of prompts or tool schemas, to enable more robust and effective tool invocation. 
% These phases work together to build memory for the tool, driving sustained improvements in the agent's performance.
These phases work together to maintain an evolving experience pool for the tool, which can be continuously updated and refined through ongoing experiences, driving sustained improvements in the agent's performance.
% an evaluator analyzes agent generated trajectories (both successful and failed)
% continuously optimizes the experience pool by incorporating new solid experiences and discarding outdated ones, ensuring stable relevance and adaptability to shifting task demands
\subsection{Experience Acquisition}
We begin with formalizing a tool invocation experience as $E=\langle h,m,c\rangle$, where $h$ is the hash-based unique identifier of the experience. The metadata $m$ captures tool invocation details, including tool information (name and schema), usage information (agent context and tool input), and execution information (tool response, success flag, time cost, and token cost). The experience content $c$ is produced by an agent $\mathrm{LLM}_{\mathrm{evaluator}}$ that takes $m$ as input and consists of: 1) $scores=[s_1,\ldots,s_{d}]$, a $d$-dimensional binary vector with $s_i\in\{0,1\}$ that characterizes different aspects of invocation quality, and 2) $explanation$, a short natural language explanation of the $scores$ that explains why the tool invocation succeeded or failed. 
% and 3) $tool\_problem$, which indicates whether the issue is caused by the tool itself rather than the agent's invocation. 
% tool contexts, inputs, or responses

 % from three different perspectives (tool context, input, and response)
% We design 10 questions (the prompt template is shown in Appendix~\ref{prompt}),
To acquire as many valuable experiences as possible for improving agents' tool use capability, {\ours} first collects trajectories in which an agent solves diverse tasks using different tools. For each agent-tool interaction in these trajectories, metadata $m$ is extracted, with a hash value $h$ computed to distinguish different invocations of the same tool under different tool context, input, or response. An agent $\mathrm{LLM}_{\mathrm{evaluator}}$ then evaluates each tool invocation to produce the experience content $c$. 
To ensure accurate and reliable evaluation, our LLM-as-a-Judge~\citep{zheng2023judging} method guides $\mathrm{LLM}_{\mathrm{evaluator}}$ to answer a series of simple yes-or-no questions, such as ``Are the input parameters in the correct format for their respective fields?''. 
The evaluation can be detailed into $d$ aspects, and the answers to the $d$ binary questions are encoded as a $d$-dimensional vector $scores$. 
By providing a natural-language explanation for each dimension, denoted as $explanation$, $\mathrm{LLM}_{\mathrm{evaluator}}$ can attribute the reasons for a tool invocation's success or failure. A generic evaluation we designed for tool invocation covers 10 aspects (the prompt template is shown in Appendix~\ref{prompt}), and the corresponding explanation provides comprehensive information by grouping these aspects from three perspectives: (1) \textbf{tool-use environment}, which assesses whether the tool is selected appropriately and invoked at the right time given the context;  (2) \textbf{agent behavior}, which assesses whether the agent invokes the tool with an input that is correct both syntactically and semantically; and (3) \textbf{tool reliability}, which assesses whether the tool executes stably and returns a response as expected. All experiences acquired are indexed by tool name and stored into a vector database, forming an initial structured experience pool for the tool. 
% Moreover, we highlight the causes of failures and instruct $\mathrm{LLM}_{\mathrm{evaluator}}$ to judge whether the error is caused by the tool, such as low availability or behavior inconsistent with its schema. 
% multiple
% \footnote{\url{https://en.wikipedia.org/wiki/Equivalence_class}}
With the notion of equivalence classes~\cite{class}, {\ours} naturally utilizes the output of the evaluation and partitions the experience pool $\mathcal{E}_\tau=\{E_i\}_{i=1}^{N_\tau}$ of the same tool $\tau$, into $K$ equivalence classes based on their $scores$, as shown in Eq.~\ref{eq1}: 
% \zhaoyang{unify the notation of scores?
% add reference of equivalence classes or explain how we define the equivalence classes}
\begin{equation}
\label{eq1}
\begin{aligned}
\kappa(E) &\triangleq scores(E)\in\{0,1\}^{d},\\
\{\mathcal{C}_k\}_{k=1}^{K} &\leftarrow \{\kappa^{-1}(v): v\in\kappa(\mathcal{E}_\tau)\}.
\end{aligned}
\end{equation}
Each equivalence class corresponds to a distinct pattern of tool invocation. For example, the perfect invocation pattern can be denoted as class $\mathcal{C}_{\mathrm{perfect}} \triangleq \kappa^{-1}\!\left([1,1,\ldots,1]\right)$, which indicates that the tool is invoked in an appropriate context with correct inputs and returns the expected response. 
% Another special class is defined as $\mathcal{C}_{\mathrm{tool\_problem}} \triangleq \kappa^{-1}\!\left([1,1,1,1,1,1,0,0,0,0]\right)$, which captures cases where the tool is invoked in an appropriate context with correct inputs, yet the response deviates substantially from expectations. In such cases, the failure is more likely to be attributable to the tool itself (e.g., low availability or an implementation inconsistent with its schema) rather than to the agent’s invocation behavior.

{
\SetAlgoSkip{1pt}      % 算法浮动体 与 上下正文 的间距（局部）
% \SetAlgoSkipEnd{2pt}   % 算法结束处额外间距（ruled 时尤其明显）
\begin{algorithm}[t]
\SetAlCapFnt{\small}
\SetAlCapNameFnt{\small}
\DontPrintSemicolon
\caption{Equivalence-Class-Based Experience Selection Algorithm for Many-shot Tool}
\label{algo:ecs_short}
\small
\KwIn{Equivalence classes $\{\mathcal{C}_k\}_{k=1}^{K}$ of tool $\tau$ derived from $\mathcal{E}_\tau$; Target selection quota $Q$}
\KwOut{Selected subset $\mathcal{S}_\tau$}

\BlankLine
{\color{lightblue}{// Select one from each of the $Q$ largest classes when the number of classes exceeds the quota}}\;
\If{$K>Q$}{
  $(\mathcal{C}_{(1)},\ldots,\mathcal{C}_{(K)}) \leftarrow \textsc{Sort}_{\mathrm{desc}}(\{\mathcal{C}_k\}_{k=1}^{K}, \,|\cdot|)$\;
  $\mathcal{S}_\tau \leftarrow \emptyset$\;
  \For{$k\leftarrow 1$ \KwTo $Q$}{
    $E \leftarrow \textsc{Sample}(\mathcal{C}_{(k)},1)$ ,  $\mathcal{S}_\tau \leftarrow \mathcal{S}_\tau \cup \{E\}$\;
  }
}
\Else{
  {\color{lightblue}{// Coarse-grained allocation}}\;
  \For{$k\leftarrow 1$ \KwTo $K$}{
    % {\color{lightblue}{// At least one sample per class}}\;
    $a_k\leftarrow 1$, 
    % {\color{lightblue}{// Proportional distribution by class sizes}}\;
    $n_k\leftarrow |\mathcal{C}_k|$, $p_k\leftarrow n_k/\sum_{j=1}^{K} n_j$\;
    $a_k\leftarrow a_k+\lfloor (Q-K)\cdot p_k\rfloor$\;
    \If{$a_k>n_k$}{
      $a_k\leftarrow n_k$\;
    }
  }

  {\color{lightblue}{// Fine-grained adjustment}}\;
  $\sigma \leftarrow \textsc{Argsort}_{\mathrm{desc}}(\{p_k\}_{k=1}^{K})$, 
  $L \leftarrow Q-\sum_{k=1}^{K} a_k$\;
  \While{$L>0$ \textbf{and} $\exists k: a_k<n_k$}{
    \For{$t\leftarrow 1$ \KwTo $K$}{
      $k\leftarrow \sigma(t)$\;
      \If{$a_k<n_k$ \textbf{and} $L>0$}{
        $a_k\leftarrow a_k+1$, $L\leftarrow L-1$\;
      }
    }
  }

  {\color{lightblue}{// In-class sampling}}\;
  $\mathcal{S}_\tau \leftarrow \emptyset$\;
  \For{$k\leftarrow 1$ \KwTo $K$}{
    $\mathcal{S}_\tau \leftarrow \mathcal{S}_\tau \cup \textsc{Sample}(\mathcal{C}_k,a_k)$\;
  }
}
\Return $\mathcal{S}_\tau$\;
\end{algorithm}
}

\subsection{Experience Distillation}\label{sec:reuse}

Equipped with rich experiences, the goal of the second phase in {\ours} is to distill various invocation-level experiences into unified tool-level guidance, which can be flexibly generalized to diverse future tasks and environments. 
Concretely, this phase consists of two steps: \textbf{experience filter} and \textbf{experience summary}. The former keeps the experience pool up-to-date and high-quality by removing unhelpful experiences and selecting helpful ones, while the latter analyzes and summarizes the retained experiences, deriving insightful and practical guidance that characterizes both qualitative tool-use behavior and quantitative tool-use performance. 

% insightful  from comparison and synthesis
% organizing the fragmented experiences into unified and reusable guidance. 

In the experience filtering step, {\ours} removes redundant tool invocation experiences for each tool based on their hash values. It then sorts the experiences by their creation time (which is correlated with the actual invocation time) and discards outdated ones to ensure the timeliness of the experience pool. Then, we perform an effectiveness check with a sliding time window to remove bad experiences that may introduce negative effects. For a given timestamp, if the average $scores$ of the next three consecutive timestamps are consistently lower than that of the current timestamp, all invocation experiences associated with the current timestamp that are used for distilling guidance are removed. 
After removing unhelpful experiences, {\ours} implements different selection strategies by categorizing tools into three groups according to the number of valid experiences: 1) \textbf{one-shot tool}: 
to avoid biased guidance summarized from only a single invocation, no experience 
is selected and the subsequent summary step is skipped for the tool; 2) \textbf{few-shot tool}: all retained experiences are used to summarize comprehensive guidance for the tool; and 3) \textbf{many-shot tool}: a designed equivalence-class-based selection method is applied to select at most $Q$ representative experiences, where $Q$ is computed according to the available context length. 
% We outline the procedure in Algorithm~\ref{algo:ecs_short}, 
The procedure is described in Algorithm~\ref{algo:ecs_short} and analyzed in depth in Appendix~\ref{proof}. 
The method promotes diversity while preserving the original score distribution, thereby ensuring that the selected experiences are representative by covering as many invocation patterns as possible while emphasizing the most frequently observed ones. 
% as they capture the tool’s typical success and failure patterns. 
% selects a maximally diverse subset while preserving the original score distribution as closely as possible
% 在经验过滤（Experience Filter）步骤，{\ours}首先根据每个工具收集的所有经验的hash移除冗余的工具调用经验。然后根据经验的创建时间（往往和工具调用时间有强相关性）进行排序，并移除过期经验，这一点保证了工具记忆的时效性。然后，我们基于时间窗口的对每条经验进行效能判断，用来移除添加后反而造成负面影响的经验。具体来说，{\ours}收集每个时间点该工具的所有经验计算平均得分，若某个时间戳后续连续3个时间戳的平均得分均低于该时间戳，则删除该时间戳下所有已总结的调用记录。接下来，我们精心设计了一种经验选择方法，该方法根据工具的有效经验数将其分成了三类：1）one-shot tool，为了避免从单条调用中获得片面的信息，{\ours}将不选择任何一条经验，使其跳过后一步的总结过程 2）few-shot tool，该工具的调用经验较少，{\ours}将利用所有经验得到全面的guidance，and 3) many-shot tool，该工具的调用经验数量多且各种各样，为此，我们设计了一种基于等价类的选择算法，选择出那些能代表工具使用成功和失败范式的经验，实现如\ref{algo}。算法的核心思想是在保证了所选经验代表性的前提下，尽量保持与原始数据质量分布的一致性，从而为后一步提供全面且准确刻画了工具的调用经验。

After experience filtering is completed, {\ours} conducts qualitative behavior analysis alongside quantitative performance analysis over the retained experiences, producing summarized guidance. The qualitative analysis reflects how the tool should be used. Prompted with chain-of-thought reasoning (see Appendix~\ref{prompt}), an agent $\mathrm{LLM}_{\mathrm{summarizer}}$ can summarize the usage behavior for the tool from four aspects: core function, success patterns, common issues, and best practices. The quantitative analysis reflects the tool's performance and is based on statistics over historical experiences, including the invocations' average success rate, average score, average time cost, and average token cost. 

\subsection{Experience Reuse}\label{sec:refinement}
% accumulated experiences 
The final phase of {\ours} is experience reuse, which aims to apply the guidance to new task scenarios, enabling the agent to shift from blind trial-and-error tool use to informed guided tool use. Instead of treating all distilled guidance equally, {\ours} categorizes it into dynamic and stable types based on the consistency of invocation patterns, and applies different usage strategies accordingly. 
%  {\ours}的最后一个阶段是经验重用，旨在将历史积累的工具使用经验有效应用于新的任务场景中，帮助智能体从试错使用演变为指导推理。 经验重用的核心创新在于一个自适应的应用策略。系统并非简单地将所有检索到的工具guidance统一处理，而是根据历史工具调用结果的统计特征，设计了两种不同的使用模式。具体而言，因为{\ours}是一个动态机制允许工具记忆和生成的guidance都随着时间环境而变化，所以默认的情况下guidance将短期的提示词的形式发挥作用，其中会以较宽松的方式引导模型以一种更智能的方式使用工具。另外一种模式则对应了经过长期积累且经过反复验证的稳定经验，这种时候生成的guidance会转化为工具配置级别的约束条件，以更严格和直接的方式指导模型行为。我们通过公式1来识别这类长期guidance，对应着工具在多次调用中展现出一致的问题模式或被多次标记为工具本身存在问题的情况。值的注意的是，即使该guidance被内化为长期的工具schema，在使用过程中也并非永久不变，随着更多新调用经验的加入，工具记忆可能会被其它使用模式占据主要空间，从而获得新的长期guidance。

Since {\ours} is an adaptive mechanism and the experience pool can be updated over time and across environments, dynamic guidance is by default used as lightweight contextual prompts to guide the agent to invoke tools appropriately. In contrast, stable guidance corresponds to reliable experiences that have been extensively validated. In this case, the distilled guidance is combined into the tool schema, constraining agent behavior more directly and strictly. The categorization criterion is given in Eq.~\ref{eq2}, where $C^{*}$ denotes the largest equivalence class among all classes except the perfect invocation class $\mathcal{C}_{\mathrm{perfect}}$ (perfect tool use does not require stricter constraints), and $\alpha$ is a threshold that controls the strictness of the criterion. Specifically, when the number of experiences reaches half of the selection quota (${Q}/{2}$) and most experiences belong to a single non-perfect class, the criterion function $f$ is triggered and the guidance is classified as stable. And $\alpha$ is recommended to be set as $K/Q$, which comes from the mixing coefficient in the experience selection strategies as analyzed in the Appendix ~\ref{proof}, capturing the trade-off between diversity coverage and distribution preservation. Using this threshold ensures that stable guidance is triggered only when a high-frequency and consistent equivalence class dominates. The stable guidance covers situations in which a tool exhibits the consistent failure pattern across different invocations. For example, a typical stable guidance may reflect tool usability issues under our generic evaluation. That is, even when the tool is invoked in an appropriate context with correct inputs, its response can still deviate substantially from expectations, causing the equivalence class $\mathcal{C} \triangleq \kappa^{-1}\!\left([1,1,1,1,1,1,0,0,0,0]\right)$ to dominate. 
\begin{equation}
\label{eq2}
% \small
\begin{gathered}
\textsc{Type}(guidance_\tau)
=
\begin{cases}
\text{stable}, & \text{if } f(\mathcal{E}_\tau)\\
\text{dynamic}, & \text{otherwise}
\end{cases}\\[4pt]
f(\mathcal{E}_\tau)
\triangleq
\Big(|\mathcal{E}_\tau|\ge \frac{Q}{2}\Big)\ \wedge\ (|\mathcal{C}^*|\ge \alpha\,|\mathcal{E}_\tau|).
\end{gathered}
\end{equation}

Overall, experience reuse leverages distilled guidance to steer tool invocation in new scenarios. The different usage strategy allows {\ours} to choose an appropriate way to apply guidance based on its reliability and generalizability. It keeps dynamic guidance flexible to adapt to changing environments, while ensuring the validated stable guidance provides consistent constraints across tasks.

% 这种自适应策略使得系统能够根据经验的成熟度和可靠性，选择最合适的知识传递方式和起作用时间，既能够保持新兴或变化中经验的灵活性以适应各种环境的动态，又能通过配置级约束确保经过充分验证的关键经验得到严格执行，更持续稳定地提升工具使用的成功率和任务整体效能。

\section{Experiments}
\subsection{Experimental Settings}

\paragraph{Baselines.} 
To evaluate the effectiveness of {\ours}, we compare it with: (1) No Method, and three different types of baseline methods improve the agent’s tool use by learning from prior experience (2) Few-shot learning~\cite{fewshot}, a prompting method that provides a small number of tool-use examples in the prompt to let the model learn from them, (3) DRAFT~\cite{Draft}, a method that dynamically improves the quality of tool documents to enhance agents’ understanding and use of tools, and (4) Mem0~\citep{chhikara2025mem0}, a method which builds conversational memory for an agent by extracting key information from its interactions with the external world. 
These three methods provide guidance for tool-use from different aspects, with further details provided in Appendix~\ref{app:baselines}. 
% (2)Dspy~\cite{dspy}, a method for improving an agent’s prompts by creating and collecting examples from previously completed tasks

\vspace{-1mm}
\paragraph{Datasets and Metrics.} % Datasets and Metrics
The task-solving process of tool-augmented LLMs is often decomposed into four stages: task planning, tool selection, tool calling, and response generation~\cite{toolsurvey}.
To thoroughly evaluate the effectiveness of {\ours}, we conduct experiments on three benchmark datasets, each targeting different stages of tool usage: MetaTool~\cite{metatool}, API-Bank~\cite{apibank}, BFCL-V3~\citep{patilberkeley}. 
For MetaTool, we select the 100 most frequently used tools and sample up to 10 tasks for each tool under each sub-task setting, resulting in 1,669 tasks in total. For API-Bank, we use the full lv1 dataset, comprising 399 tasks. Both datasets are evaluated using leave-one-out cross-validation. For BFCL-V3, since some tasks involve tools that are never reused in other tasks, we filter out these cases and retain 461 tasks for the test set and 1542 tasks for the training set. 
We use both Avg@3 and Pass@3 as evaluation metrics. Avg@3 denotes the average task success rate over three independent runs, while Pass@3 measures the probability that at least one of the three runs is successful. As different datasets target different stages of tool usage, we follow their original evaluation criteria when defining ``successful'' to maintain consistency. See Appendix~\ref{app:dataset} for more dataset and metric details.

\vspace{-1mm}
\paragraph{Implementation Details.} 
We conduct experiments on GPT-5 nano~\cite{openai2025gpt5}, Deepseek-V3~\cite{deepseekai2025deepseekv3technicalreport}, and the Qwen3 series models~\citep{qwen3technicalreport} to demonstrate the effectiveness of {\ours} across different model architectures and scales. We use prompt-based tool invocation method and set $temperature =0.5$ for trajectory sampling. 
During experience acquisition and experience distillation, we adopt Qwen3-max as the $\mathrm{LLM}_{\mathrm{evaluator}}$ and $\mathrm{LLM}_{\mathrm{summarizer}}$ to produce reliable tool invocation experiences and distilled guidance. 
During experience reuse, we set $\alpha=0.8$ to control the criterion in our different usage strategy of guidance. 
In addition, to support our design of removing unhelpful experiences over time, we execute tasks sequentially. 
For fair comparison, we keep these settings fixed unless otherwise specified in ablation studies. 
% including the extraction process of metadata $m$ in the experiences， the automatic calculation method for the hyperparameter quota $Q$, and other things we tried
More implementation details can be found in the Appendix~\ref{app:tried}.

\begin{table*}[t]
    \caption{Performance comparison (\%) across MetaTool, API-Bank, and BFCL-V3. \textbf{Bold} indicates the best results within each model.}
    \centering
    \resizebox{1\linewidth}{!}{
    \begin{tabular}{llcccccccccccc}
    \toprule
         \multirow{2}{*}{\textbf{Model}} & \multirow{2}{*}{\textbf{Method}} &
         \multicolumn{3}{c}{\textbf{MetaTool}} &
         \multicolumn{3}{c}{\textbf{API-Bank}} &
         \multicolumn{3}{c}{\textbf{BFCL-V3}} &
         \multicolumn{3}{c}{\textbf{Total}} \\
         &  & Pass@1 & Avg@3 & Pass@3 & Pass@1 & Avg@3 & Pass@3 & Pass@1 & Avg@3 & Pass@3 & Pass@1 & Avg@3 & Pass@3 \\
    \midrule
    \multirow{5}{*}{\textbf{GPT-5 nano}}
        & No Method & 72.62 & 72.76 & 78.49 & 82.96 & 83.46 & 86.97 & 53.80 & 53.00 & 60.95 & 70.82 & 70.62 & 76.63 \\
        & Few-shot  & 74.12 & 75.11 & 82.32 & 83.71 & 83.96 & \textbf{87.22} & 56.18 & 55.24 & 61.39 & 72.36 & 72.65 & 79.28 \\
        & DRAFT     & 73.94 & 73.04 & 78.97 & 84.21 & 83.46 & \textbf{87.22} & 57.27 & 57.27 & 62.26 & 72.52 & 71.58 & 77.23 \\
        & Mem0      & 74.96 & 76.13 & 82.92 & 84.96 & 85.21 & \textbf{87.22} & 60.95 & 61.61 & 65.08 & 73.98 & 74.67 & 80.35 \\
        & \cellcolor{gray!15}{\ours} & \cellcolor{gray!15}\textbf{81.67} & \cellcolor{gray!15}\textbf{82.07} & \cellcolor{gray!15}\textbf{84.60} & \cellcolor{gray!15}\textbf{86.72} & \cellcolor{gray!15}\textbf{86.55} & \cellcolor{gray!15}\textbf{87.22} & \cellcolor{gray!15}\textbf{64.43} & \cellcolor{gray!15}\textbf{63.99} & \cellcolor{gray!15}\textbf{66.38} & \cellcolor{gray!15}\textbf{79.32} & \cellcolor{gray!15}\textbf{79.22} & \cellcolor{gray!15}\textbf{81.69} \\
    \midrule
    \multirow{5}{*}{\textbf{DeepSeek-V3}}
        & No Method & 83.10 & 82.94 & 84.66 & 84.71 & 84.38 & 85.46 & 58.79 & 59.65 & 65.94 & 78.92 & 78.66 & 81.37 \\
        & Few-shot  & 82.74 & 83.90 & 86.28 & 85.21 & 84.63 & 86.22 & 60.52 & 60.30 & 67.90 & 79.08 & 79.45 & 82.92 \\
        & DRAFT     & 80.23 & 80.79 & 82.44 & 84.96 & 85.63 & 86.47 & 62.26 & 61.61 & 68.55 & 77.70 & 77.80 & 80.54 \\
        & Mem0      & 83.88 & 84.56 & 86.40 & 85.46 & 85.55 & 86.47 & 65.08 & 65.15 & 68.33 & 80.70 & 80.91 & 83.12 \\
        & \cellcolor{gray!15}{\ours} & \cellcolor{gray!15}\textbf{85.26} & \cellcolor{gray!15}\textbf{85.38} & \cellcolor{gray!15}\textbf{86.52} & \cellcolor{gray!15}\textbf{87.72} & \cellcolor{gray!15}\textbf{87.39} & \cellcolor{gray!15}\textbf{87.97} & \cellcolor{gray!15}\textbf{69.41} & \cellcolor{gray!15}\textbf{69.92} & \cellcolor{gray!15}\textbf{72.02} & \cellcolor{gray!15}\textbf{82.76} & \cellcolor{gray!15}\textbf{82.61} & \cellcolor{gray!15}\textbf{84.11} \\
    \midrule
    \multirow{5}{*}{\textbf{Qwen3-8B}}
        & No Method & 76.51 & 76.97 & 77.71 & 83.96 & 83.88 & 84.21 & 58.79 & 58.28 & 60.30 & 74.46 & 74.41 & 75.56 \\
        & Few-shot  & 79.93 & 79.83 & 82.92 & 83.71 & 82.62 & 84.96 & 60.09 & 59.29 & 61.39 & 76.91 & 76.27 & 79.32 \\
        & DRAFT     & 78.19 & 77.33 & 77.89 & 85.71 & 84.96 & 85.46 & 60.74 & 60.30 & 62.91 & 76.20 & 75.18 & 76.35 \\
        & Mem0      & 75.07 & 75.47 & 82.38 & 86.22 & 86.05 & 86.47 & 63.34 & 64.93 & 66.16 & 74.69 & 74.98 & 80.07 \\
        & \cellcolor{gray!15}{\ours} & \cellcolor{gray!15}\textbf{83.52} & \cellcolor{gray!15}\textbf{84.88} & \cellcolor{gray!15}\textbf{85.08} & \cellcolor{gray!15}\textbf{86.47} & \cellcolor{gray!15}\textbf{87.89} & \cellcolor{gray!15}\textbf{87.97} & \cellcolor{gray!15}\textbf{67.46} & \cellcolor{gray!15}\textbf{66.96} & \cellcolor{gray!15}\textbf{68.33} & \cellcolor{gray!15}\textbf{81.06} & \cellcolor{gray!15}\textbf{81.82} & \cellcolor{gray!15}\textbf{82.48} \\
    \midrule
    \multirow{2}{*}{\textbf{Qwen3-32B}}
        & No Method & 80.05 & 79.43 & 80.17 & 84.71 & 84.88 & 85.21 & 65.15 & 65.08 & 66.16 & 78.05 & 77.55 & 78.41 \\
        & \cellcolor{gray!15}{\ours} & \cellcolor{gray!15}\textbf{84.68} & \cellcolor{gray!15}\textbf{85.02} & \cellcolor{gray!15}\textbf{86.28} & \cellcolor{gray!15}\textbf{86.97} & \cellcolor{gray!15}\textbf{87.30} & \cellcolor{gray!15}\textbf{87.72} & \cellcolor{gray!15}\textbf{70.72} & \cellcolor{gray!15}\textbf{71.01} & \cellcolor{gray!15}\textbf{73.32} & \cellcolor{gray!15}\textbf{82.48} & \cellcolor{gray!15}\textbf{82.56} & \cellcolor{gray!15}\textbf{84.14} \\
    \midrule
    \multirow{2}{*}{\textbf{Qwen3-235B}}
        & No Method & 78.25 & 79.23 & 80.29 & 85.46 & 85.46 & 85.71 & 71.37 & 71.15 & 73.54 & 78.13 & 78.49 & 79.91 \\
        & \cellcolor{gray!15}{\ours} & \cellcolor{gray!15}\textbf{86.34} & \cellcolor{gray!15}\textbf{86.70} & \cellcolor{gray!15}\textbf{86.94} & \cellcolor{gray!15}\textbf{87.47} & \cellcolor{gray!15}\textbf{86.97} & \cellcolor{gray!15}\textbf{88.22} & \cellcolor{gray!15}\textbf{79.61} & \cellcolor{gray!15}\textbf{78.52} & \cellcolor{gray!15}\textbf{80.04} & \cellcolor{gray!15}\textbf{85.29} & \cellcolor{gray!15}\textbf{84.98} & \cellcolor{gray!15}\textbf{85.69} \\
    \bottomrule
    \end{tabular}
    }
    \label{tab:main}
\vspace{-1mm}
\end{table*}

\subsection{Main Results} 
% 表~\ref{tab:results} 汇总了五种骨干模型在 MetaTool、API-Bank 与 BFCL-V3 上的实验结果。在每个模型组中，{\ours} 都在三项基准的 Avg@3 与 Pass@3 上取得最优成绩，从而带来sota的表现。以 Qwen3-8B 为例，相比 No Method，{\ours} 在Total 指标上分别带来+7.41 的 Avg@3 与 +6.92 的 Pass@3，且这种增益可以持续到更大规模的模型变体上，验证了我们方法的通用性。值得注意的是，配备 {\ours} 的小模型能够达到与更大模型（但不使用记忆）相当甚至更优的表现，这表明，有效的记忆机制能够显著缩小不同模型规模之间的性能差距。
Table~\ref{tab:main} reports the main results of {\ours} across models on three datasets that target different phases of tool invocation. 
Overall, {\ours} achieves the best Avg@3 and Pass@3 on every dataset, consistently outperforming all competing methods.   
For Qwen3-8B, {\ours} improves the total metric by
{\setlength{\fboxsep}{0.5pt}\colorbox{gainbg}{7.41$\uparrow$}} in Avg@3 and
{\setlength{\fboxsep}{0.5pt}\colorbox{gainbg}{6.92$\uparrow$}} in Pass@3 over the ``No Method'' setting. We also report Pass@1, which highlights clearer distinctions between methods under lower inference budgets. 
These gains persist as model size increases, demonstrating the generality of our approach. 
Importantly, {\ours} allows smaller models to match or even surpass larger models in the ``No Method'' setting, indicating that our approach helps mitigate the tool-use capability gap between weaker and stronger models. 

% \begin{figure}[t]
%     \centering
%     \includegraphics[width=0.5\linewidth]{latex/figures/output.png}
%     \caption{Avg@3 performance (\%) on MetaTool and BFCL-V3 across different sub-task settings.}
%     \vspace{-2mm}
%     \label{fig:detail}
% \end{figure}

\begin{figure}[t]
    \centering
    \begin{minipage}{0.44\linewidth}
        \centering
        \includegraphics[width=\linewidth]{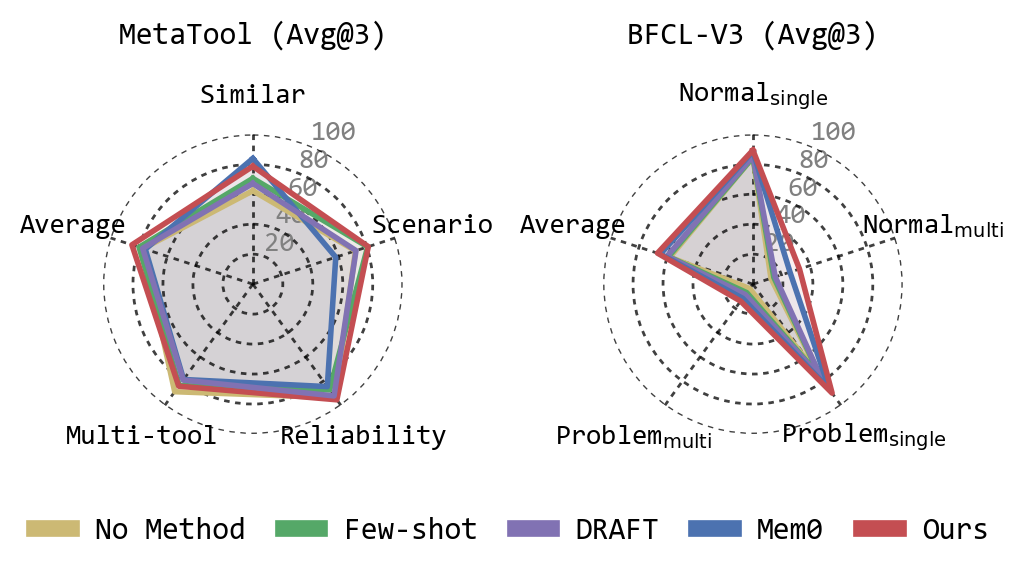}
        \caption{Avg@3 performance (\%) on MetaTool and BFCL-V3 across sub-task settings.}
        \label{fig:detail}
    \end{minipage}\hfill
    \begin{minipage}{0.53\linewidth}
        \centering
        \includegraphics[width=\linewidth]{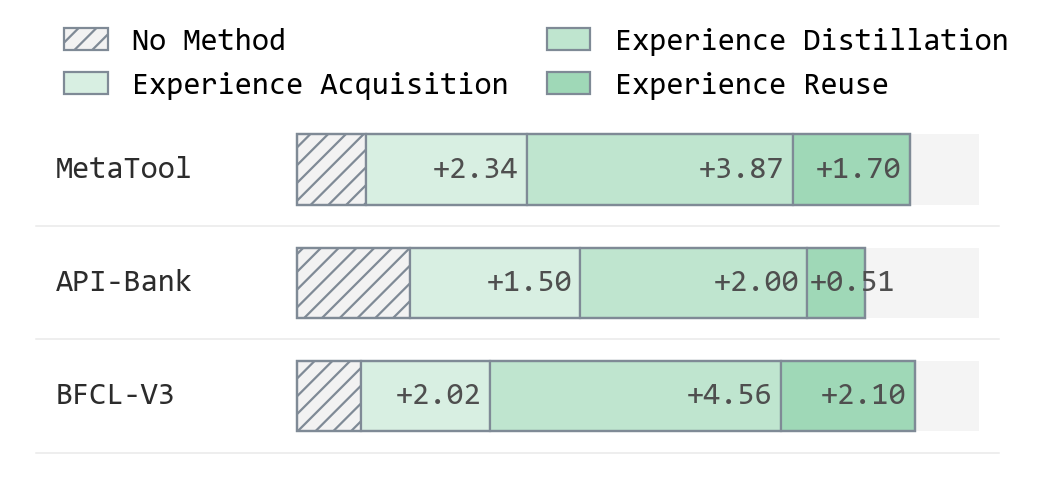}
        \caption{Ablation results (Avg@3, \%) of different phases in {\ours}.}
        \label{fig:ablation}
    \end{minipage}
\end{figure}

% 除总体表现外，我们还对每个基准进行了深入分析，图1给出了在MetaTool和BFCL-v3数据集中的子任务上的表现。具体来说，工具选择（MetaTool）任务中，{\ours} 在区分相似工具和工具存在可靠性问题时格外突出。工具调用（API-Bank）任务中，我们观察到了不同模型在几个方法上的相似指标，并推测是由于该数据集设定相对简单，到达了针对工具调用方面的优化上限所造成的。在BFCL-v3上，我们的方法在问题设定（缺失参数，缺失方法，不相关方法等）子集上的表现非常亮眼，这表明{\ours}从过去经验中学习到了如何正确面对这些不完美情况。此外，一个有趣的统计是我们的方法在常规的多轮任务中平均推理步骤从10.14减少至9.02，而针对有问题的多轮任务则增加了2.4步。这表明正常推理时{\ours}帮助模型在较少次数的尝试内提高成功概率，而面对不完美情况时候，模型有意识通过和用户交互或提前检查环境去解决这些问题。
Beyond the overall performance, we further conduct a detailed analysis of {\ours} on each dataset. 
Figure~\ref{fig:detail} presents the results on MetaTool and BFCL-V3 across different sub-task settings. 
In tool selection (MetaTool) tasks, {\ours} is particularly effective ({\setlength{\fboxsep}{0.5pt}\colorbox{gainbg}{16.34$\uparrow$}}) when faced with distractor tools that are very similar. 
% or have reliability issues({\setlength{\fboxsep}{0.5pt}\colorbox{gainbg}{6.20$\uparrow$}}). the model is
In tool calling (API-Bank) tasks, we observe that different methods achieve comparable scores. 
% tool-augmented
This is likely because the dataset is relatively simplistic, allowing all approaches to bring model performance close to the upper limit of their inherent capabilities. 
In response generation tasks (BFCL-V3), {\ours} demonstrates strong performance ({\setlength{\fboxsep}{0.5pt}\colorbox{gainbg}{13.09$\uparrow$}} in single-turn tasks and {\setlength{\fboxsep}{0.5pt}\colorbox{gainbg}{10.33$\uparrow$}} in multi-turn tasks) on noisy settings, such as missing parameters, undefined tools or irrelevant tool choices. 
% on imperfect information settings
% This indicates that {\ours} helps the agent generalize tool use effectively in unpredictable situations. 
This indicates that {\ours} helps the agent generalize tool use effectively in unpredictable or even problematic environments.
% This indicates {\ours} help agent successfully generalizes the tool usage to imperfect scenarios. problematic
% This suggests that the agent has learned to handle such imperfect scenarios based on prior experience. 
Besides, we observe an interesting phenomenon that our method reduces the average inference steps in normal multi-turn tasks from 10.14 to 9.02, but increases the step count by 2.4 in problematic multi-turn tasks. 
% This indicates that {\ours} enables the model to achieve higher success rates with fewer trials under normal settings, while encouraging more deliberate behaviors when facing challenging scenarios. 
This indicates that {\ours} enables the agent to achieve higher success rates with fewer trials under normal settings, reaching correct solutions with fewer unnecessary interactions. At the same time, it encourages more deliberate behaviors in challenging scenarios, prompting the agent to interact with the environment to ensure more reliable tool use. 
Such a shift toward a more flexible tool-use strategy is desirable, as it suggests that the agent becomes more robust across diverse situations under our guidance.
% by consciously interacting with users or conducting anticipatory environmental checks to address potential problems.

% \begin{figure}[t]
%     \centering
%     \includegraphics[width=0.5\linewidth]{latex/figures/output.png}
%     \caption{Avg@3 performance (\%) on MetaTool and BFCL-V3 across different sub-task settings.}
%     \vspace{-2mm}
%     \label{fig:detail}
% \end{figure}

\subsection{Ablation Studies}\label{sec:ablation}
As shown in Figure~\ref{fig:ablation}, we conduct an ablation study of the key components of {\ours} using Qwen3-8B as the backbone model. Compared with the baseline without any method, progressively adding each phase yields consistent improvements. First, only using \textit{experience acquisition} as guidance during inference achieves better performance than few-shot learning, indicating that structured experiences acquired from task interactions provide more effective signals. Second, adding \textit{experience distillation} results in a substantial additional gain, highlighting the importance of maintaining a high-quality experience pool and aggregating multiple experiences. Finally, although the additional improvement after adding \textit{experience reuse} is milder than that of the first two phases, it is crucial for driving long-term evolution of the agent and tool quality in real applications.

\begin{table}[tbp]
\caption{Effect of the number of acquired experiences on MetaTool performance (Avg@3, \%) in {\ours}. $\Delta$ denotes the absolute change compared to the previous setting.}
\centering
\resizebox{0.7\linewidth}{!}{%
    \begin{tabular}{c|cccccccc}
    \toprule
    \textbf{\# Number} & 0 & 3 & 6 & 9 & 11 & 14 & 17 & 20 \\
    \midrule
    \textbf{Performance (\%)} & 77.29 & 82.26 & 85.80 & 86.70 & 86.16 & 84.42 & 85.20 & 86.22 \\
    \textbf{$\Delta$ (\%)} & -- & +4.97 & +3.54 & +0.90 & $-0.54$ & $-1.74$ & +0.78 & +1.02 \\
    \bottomrule
    \end{tabular}%
}
\label{tab:experience_num}
\vspace{-2mm}
\end{table}

Additionally, to study how the number of experiences affects {\ours}, we conduct experiments on MetaTool, selecting tools with sufficiently diverse invocations. Table~\ref{tab:experience_num} shows that {\ours} improves quickly once the number of experiences for a tool meets our few-shot tool criterion. As more experiences are acquired, performance peaks at around 9 experiences, after which adding additional experiences introduces fluctuations. With further increases in experience, {\ours} stabilizes and maintains strong performance, as the experience filtering phase continuously removes unhelpful experiences while keeping the most helpful ones.

\subsection{More Analysis}\label{sec:moreana}
\begin{table}[tbp]
\centering
\begin{minipage}{0.48\linewidth}
\centering
\caption{Impact of agent size on BFCL-V3 performance (\%) in {\ours}. $\triangle$ indicates the change relative to the original Qwen3-max setting.}
\footnotesize
\setlength\tabcolsep{3pt}
\resizebox{\linewidth}{!}{%
    \begin{tabular}{c|c|cc}
    \toprule
    \multirow{2}{*}{\textbf{${\text{LLM}}_{\text{evaluator}}$}} &
    \multirow{2}{*}{\textbf{${\text{LLM}}_{\text{summarizer}}$}} &
    \multicolumn{2}{c}{\textbf{BFCL-V3}}\\
    && Avg@3 (\%)& Pass@3 (\%)\\
    \midrule
    \multirow{2}{*}{Qwen3-max} & Qwen3-8B & 63.20\colorbox{pink!20}{$\textcolor{darkred}{\triangle=3.76 \downarrow}$} & 65.72\colorbox{pink!20} {$\textcolor{darkred}{\triangle=2.61 \downarrow}$} \\
    &Qwen3-32B & 65.51 \colorbox{pink!20}{$\textcolor{darkred}{\triangle=1.45 \downarrow}$} & 66.02 \colorbox{pink!20}{$\textcolor{darkred}{\triangle=2.31 \downarrow}$} \\
    \midrule
    Qwen3-8B & \multirow{2}{*}{Qwen3-max} &
    62.18\colorbox{pink!20}{$\textcolor{darkred}{\triangle=4.78 \downarrow}$} & 64.71\colorbox{pink!20} {$\textcolor{darkred}{\triangle=3.62 \downarrow}$} \\
    Qwen3-32B &  &
    64.78\colorbox{pink!20}{$\textcolor{darkred}{\triangle=2.18 \downarrow}$} &
    66.81\colorbox{pink!20}{$\textcolor{darkred}{\triangle=1.52 \downarrow}$} \\
    \midrule
    Qwen3-max & Qwen3-max &66.96 & 68.33 \\
    \bottomrule
    \end{tabular}%
}
\label{tab:summarizer}
\end{minipage}\hfill
\begin{minipage}{0.50\linewidth}
\centering
\caption{Self-evolving {\ours} with Qwen3-8B (\%). $\triangle$ indicates the change relative to No Method.}
\footnotesize
\setlength\tabcolsep{2pt}
\resizebox{\linewidth}{!}{%
    \begin{tabular}{cc|ccc}
    \toprule
    \textbf{Dataset} & \textbf{Metric} & \textbf{No Method} & \textbf{{\ours} (Qwen3-8B)} & \textbf{{\ours} (Qwen3-Max)} \\
    \midrule
    \multirow{2}{*}{MetaTool} & Avg@3 & 76.97 & 81.91\colorbox{gainbg}{$\textcolor{darkgreen}{\triangle=4.94 \uparrow}$} & 84.88\colorbox{gainbg}{$\textcolor{darkgreen}{\triangle=7.91 \uparrow}$} \\
    & Pass@3 & 77.71 & 82.03\colorbox{gainbg}{$\textcolor{darkgreen}{\triangle=4.32 \uparrow}$} & 85.08\colorbox{gainbg}{$\textcolor{darkgreen}{\triangle=7.37 \uparrow}$} \\
    \midrule
    \multirow{2}{*}{API-Bank} & Avg@3 & 83.88 & 85.88\colorbox{gainbg}{$\textcolor{darkgreen}{\triangle=2.00 \uparrow}$} & 87.89\colorbox{gainbg}{$\textcolor{darkgreen}{\triangle=4.01 \uparrow}$} \\
    & Pass@3 & 84.21 & 85.96\colorbox{gainbg}{$\textcolor{darkgreen}{\triangle=1.75 \uparrow}$} & 87.97\colorbox{gainbg}{$\textcolor{darkgreen}{\triangle=3.76 \uparrow}$} \\
    \midrule
    \multirow{2}{*}{BFCL-V3} & Avg@3 & 58.28 & 62.18\colorbox{gainbg}{$\textcolor{darkgreen}{\triangle=3.90 \uparrow}$} & 66.96\colorbox{gainbg}{$\textcolor{darkgreen}{\triangle=8.68 \uparrow}$} \\
    & Pass@3 & 60.30 & 63.99\colorbox{gainbg}{$\textcolor{darkgreen}{\triangle=3.69 \uparrow}$} & 68.33\colorbox{gainbg}{$\textcolor{darkgreen}{\triangle=8.03 \uparrow}$} \\
    \bottomrule
    \end{tabular}%
}
\label{tab:same_scale}
\end{minipage}
\end{table}

\begin{figure*}[tb]
    \centering
    \includegraphics[width=0.86\linewidth]{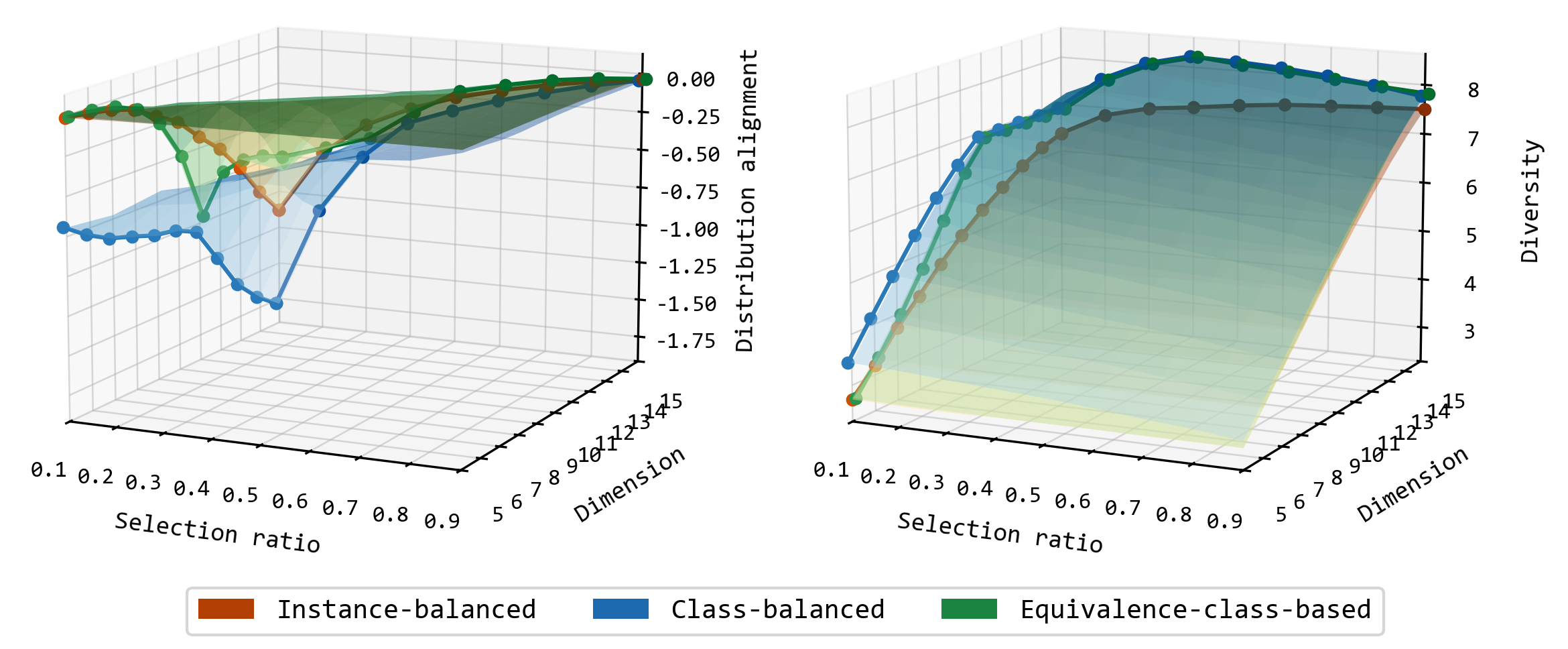}
    \caption{Distribution alignment and diversity under different selection methods.}
    \label{fig:selection}
    \vspace{-3mm}
\end{figure*}

\paragraph{Agent Capabilities.} Since {\ours} relies on the LLM-as-judge method to acquire experiences and summarize guidance effectively, we examine the impact of agent size for both roles, ${\text{LLM}}_{\text{evaluator}}$ and ${\text{LLM}}_{\text{summarizer}}$. The experiments are conducted on BFCL-V3, and the results are shown in Table~\ref{tab:summarizer}. Replacing either agent with a smaller model degrades the final task performance, with a more pronounced drop observed for ${\text{LLM}}_{\text{evaluator}}$. Specifically, substituting the ${\text{LLM}}_{\text{evaluator}}$ with the weaker Qwen3-8B yields a larger performance decline, likely because processing complex trajectories and making accurate attributions is challenging for weaker agents. In contrast, using a moderately sized model such as Qwen3-32B leads to a relatively smaller degradation.
To further check whether the gains mainly come from a high-tier model, we replace both the evaluator and the summarizer with Qwen3-8B. Table~\ref{tab:same_scale} shows that even without a high-tier model, {\ours} still yields large and stable gains over No Method on all three datasets, and these gains remain comparable to those of Qwen3-Max. This indicates that the gains mainly come from structured distillation and reuse of tool experiences, and that agents can self-evolve through {\ours} without a stronger external model.

\vspace{-2mm}
\paragraph{Equivalence-Class-Based Selection Method.} We compare our equivalence-class-based experience selection method with two typical selection methods: instance-balanced, which selects experiences uniformly, and class-balanced, which selects similar numbers from each class. We use negative Kullback-Leibler divergence~\cite{kl} to measure distribution alignment with the original experience distribution, and Shannon entropy~\cite{entropy} to measure the diversity of the selected set. Figure~\ref{fig:selection} shows the results under different dimensions $d$ of the $scores$ and selection ratios. In terms of distribution alignment, our method performs similarly to the instance-balanced method, maintaining strong consistency with the original distribution. It can even outperform the instance-balanced method through fine-grained adjustment of the selection quotas at higher selection ratios. From the perspective of diversity, the class-balanced method usually achieves the highest entropy by equalizing class proportions, whereas the instance-balanced method has lower entropy, indicating weaker class coverage. Our method achieves much higher entropy than the instance-balanced method and slightly lower entropy than the class-balanced method in most settings. It ensures that the chosen experiences both cover as many invocation patterns as possible and remain representative of the original distribution.

% 我们将经验选择步骤中基于等价类的选择方法其与两个经典选择方法进行对比。Instance-balanced方法对每个样本一视同仁、按实例均匀无放回采样。Class-balanced方法让每个类别被选中的次数尽量相近。图~\ref{fig:selection} illustrates how the three methods behave across different dimensions $d$ in $scores$ and selection ratios. 我们使用负kl来衡量其与原始经验分布的distribution alignment，使用entropy来衡量选择后经验集合的diversity。在一致性（distribution alignment）角度来看，我们的方法有着与 Instance-balanced 相近的表现，保持着与原始分布较高的一致性。此外，在高选择比例中，因其第二阶段细粒度的调整，甚至得到了比Instance-balanced更优的表现。从多样性（diversity）角度来看，Class-balanced 在通常获得最高的 entropy，因为它强调最大化类别覆盖。 Instance-balanced则体现出较低的值，说明它对稀有模式的经验覆盖不足。我们的Equivalence-class-based 方法在几乎所有设置下显著高于 Instance-balanced 而略低于 Class-balanced方法。它通过在等价类层面进行控制，从而确保所选择的经验既覆盖了尽可能多的调用模式同时又具有原始分布代表性。

% 有效提升了所选经验的模式多样性，体现出在保持分布代表性的前提下显著增强多样性的优势。
% 在强调等价类覆盖的同时，并没有显著破坏原始分布的一致性，
% Instance-balanced始终保持着接近原始经验分布的分布while Class-balanced因其拉平各个类别的概率表现始终较差。
% ，说明其在强调等价类覆盖的同时，并没有显著破坏原始分布的一致性
% Equivalence-class-based（我们的方法）：先在等价类层面建模，再在类内按实例采样，在保持整体分布对齐的同时，有针对性地覆盖更多代表性等价类，在分布对齐和多样性之间取得更好的平衡。
% 本文概述了算法\ref{algo:ecs_short}中的过程，在保留原始分数分布的同时促进多样性，通过覆盖尽可能多的调用模式同时强调最频繁观察到的模式，从而确保所选择的经验具有代表性。
\vspace{-2mm}
\paragraph{Case Analysis.} We perform case analysis to better understand the strengths and limitations of {\ours}. These cases show that {\ours} mainly helps the agent: 1) improve tool-use awareness, 2) distinguish similar tools, 3) invoke tools correctly, 4) fix tool-use errors, 5) identify tool dependencies, and 6) understand tool reliability. We also study cases where {\ours} fails, such as unseen situations, hard-to-analyze trajectories, incorrect attributions, and over-reaction. Detailed analyses and a case-coverage comparison with other methods can be found in Appendix~\ref{case}. 

\section{Conclusion}
% generalizable capability boundaries and best practices
% This paper aims to improve the robustness of tool-augmented LLM agents in unpredictable real-world settings.
This paper addresses the robustness challenges of tool use within the agent harness during runtime execution. We propose a mechanism {\ours}, which forms adaptive guidance for each tool through three phases: experience acquisition, distillation, and reuse. Extensive experiments validate the effectiveness of our method across various benchmarks, demonstrating that our guidance helps agents shift to a more informed tool-use strategy. We believe {\ours} represents an early step toward trainable tools, paving the way for building more robust and adaptive agent systems.
% We believe {\ours} represents an early step toward trainable tools, paving the way for building continuously self-evolving agent systems in the future.

\bibliography{example_paper}
\bibliographystyle{unsrt}
%%%%%%%%%%%%%%%%%%%%%%%%%%%%%%%%%%%%%%%%%%%%%%%%%%%%%%%%%%%%

\newpage
\appendix

\section{Template Prompt for LLM-as-judge Agent}\label{prompt}
\vspace{-3mm}
\begin{tcolorbox}[title={Prompt for agent ${\text{LLM}}_{\text{evaluator}}$},breakable,]
You are an expert in evaluating tool invocation process. The tool is invoked by an AI agent.\\

\textbf{Tool invocation Information:}\\
$\bullet$ Tool Name: \{tool\_name\}\\
$\bullet$ Success Flag: \{success\_flag\}\\
$\bullet$ Time Cost: \{time\_cost\}s\\
$\bullet$ Token Cost: \{token\_cost\} tokens\\
$\bullet$ Agent Context: \{context\}\\
$\bullet$ Input Parameters: \{input\_params\}\\
$\bullet$ Tool Response: \{response\}\\
$\bullet$ Tool Schema: \{schema\}\\

\textbf{Evaluation Method:} Start from a default score list of scores = [0, 0, 0, 0, 0, 0, 0, 0, 0, 0]. For each item below that is satisfied, assign 1 point to the corresponding index. The final scores should be a list of 10 integers, each being either 0 or 1.\\

\textbf{1. Use Quality (total 2 points. If context is provided, use it as an aid when evaluating):}\\
$\bullet$ Index 1: Should the tool be invoked now? Consider whether the all necessary information for the tool's invocation is ready, and whether the tool execution environment is correct. If it is a multi-round conversation, also consider the dependency relationships of the tool chain.\\
$\bullet$ Index 2: If should, is the chosen tool appropriate?\\

\textbf{2. Input Quality (total 4 points. When evaluating, consider both the context and the tool schema):}\\
$\bullet$ Index 3: Are all required parameters provided?\\
$\bullet$ Index 4: Are the input parameters valid and supported by the tool?\\
$\bullet$ Index 5: Are the input parameters in the correct format for their respective fields?\\
$\bullet$ Index 6: Does the value(content) of input parameter correctly reflect and match the given context?\\

\textbf{3. Response Quality (total 4 points):}\\
$\bullet$ Index 7: Does the response provide meaningful and useful information? or are there any error messages or information can be used as guidance for agent invoking tool better?\\
$\bullet$ Index 8: Does the response match the tool's intended purpose/function?\\
$\bullet$ Index 9: Does the response value correct (content appropriate) given the input parameters?\\
$\bullet$ Index 10: Does the response help accomplish the task within the given context?\\

\textbf{Important:}\\
1. Sometimes there is not enough information in the context or schema to make a complete evaluation. In such cases, make your best judgment based on the available information.\\
2. Some tools (commonly system tools such as mkdir, touch, echo, etc.) modify the external environment. Since these results cannot be obtained, they return ``None'' as the response. At this point, all the scores in the quality of the response should be obtained and should not be seen as a problem for the tool.\\
3. Evaluation independently from the success flag. The \texttt{success\_flag} indicates whether the tool executed without technical errors. The \texttt{evaluation} should evaluate the quality of the tool invocation. A tool can execute successfully (Success Flag=1) but still produce low-quality or irrelevant responses, leading to a low evaluation score.\\
4. Sometimes an agent will execute multiple steps and invoke multiple tools to complete a task, but you only need to evaluate the use of one tool for one of the steps, not whether the final task is completed or not.\\

\textbf{Answer Format:}\\
Please provide your answer in the following JSON format:\\

\verb|```|json\\
\{\\
\quad ``scores'': A list of 10 integers (0 or 1) based on the evaluation criteria above.\\ 
\quad ``explanation'': A brief evaluation (2-3 sentences) explaining the quality of the tool invocation, based on your evaluation. Low-quality aspects need to be reified, especially the causes of tool invocation errors.\\
\}\\
\verb|```|
\end{tcolorbox}

\begin{tcolorbox}[title={Prompt for agent ${\text{LLM}}_{\text{summarizer}}$},breakable,]
You are an expert in analyzing tool usage patterns and generating practical usage guidance for agents.\\

\textbf{Tool Information:}\\
$\bullet$ Tool Name: \{tool\_name\}\\
$\bullet$ Tool Schema: \{tool\_schema\}\\

\textbf{Recent Tool Invocation Experiences:}\\
\{experiences\}\\

\textbf{Important:}\\
1. Assume the tool (tool schema) can't be changed, your task is to guide agent to use it better.\\
2. Your answer must be based on the information given, don't make it up. If not enough data, state ``Not enough data to determine Core Function/Success Patterns/Common Issues/Best Practices.''\\
3. Your answer will be used to guide the use of the tool in the future, so do not include content related to recent tool invocation experience such as ``case \#3'' or ``Call \#2'', but some values can be used as examples.\\
4. Pay attention to information not mentioned in the tool schema, such as the response upon successful tool invocation. It's also welcome to uncover insights, such as how tools can be used more effectively, and possible dependencies between tools. But if they aren't, don't make them up.\\
5. Finally, to avoid deriving incorrect guidance from individual invocation, check whether, if the agent follows the proposed guidance, it can perform better on all recent invocation histories. If not, revise the guidance until it can. Specifically:\\
\quad$\bullet$ don't write guidance in an absolute tone without a very deterministic message (meaning that all invocation histories are satisfied, otherwise it will result in failure).\\
\quad$\bullet$ sometimes there may be inconsistencies. Consider whether this is due to the context in which the tool is being used.\\

\textbf{Your Task:}\\
Based on the tool invocation history, generate a concise and logical tool usage guidance following this structure:\\
1. \textbf{Core Function}: What this tool does and when to use it.\\
2. \textbf{Success Patterns}: Parameter patterns and usage scenarios that work well.\\
3. \textbf{Common Issues}: Main pitfalls to avoid and why they fail.\\
4. \textbf{Best Practices}: 2-3 actionable recommendations.\\

\textbf{Answer Format:}\\
Provide a structured, concise guidance (max 200 words). Focus on actionable insights derived from actual usage data. Avoid generic advice and think step by step.\\

\verb|```|txt\\
Your concise, data-driven tool usage guidance\\
\verb|```|
\end{tcolorbox}

% \textcolor{red}{Please skip this section while reading it and revise it after we decided on the algorithm.}
\section{Analysis of Equivalence-Class-Based Selection Method}\label{proof}
\subsection{Problem Restatement and Notation}
We consider a budgeted experience selection problem for a fixed tool $\tau$.
Let $\mathcal{E}_\tau=\{E_i\}_{i=1}^{N_\tau}$ be the experience pool, where
$N_\tau \triangleq |\mathcal{E}_\tau|$.
We select $Q$ experiences under a context-length budget. 
In our structured definition, each experience $E$ is associated with a binary score vector
\begin{equation}
\kappa(E)\triangleq scores(E)\in\{0,1\}^d,
\end{equation}
where $d$ is the score dimension.
Experiences are grouped into equivalence classes according to identical score vectors.
Let $\{\mathcal{C}_k\}_{k=1}^{K}$ denote the induced partition of $\mathcal{E}_\tau$, and let $K$ be the number of
observed equivalence classes.
The class distribution on the experience pool $\mathcal{E}_\tau$ is $p_k \triangleq {|\mathcal{C}_k|}/{N_\tau}$ for $k=1,\dots,K$, and $p_k>0$.  
After selection, we obtain a distribution $q=(q_1,\dots,q_K)$ in the selected set over the same $K$ classes,
where $q_k$ denotes the fraction of selected experiences belonging to class $\mathcal{C}_k$ and
$\sum_{k=1}^K q_k = 1$.

Our selection method can be viewed at the class-distribution level as forming a convex combination between
the uniform distribution (encouraging coverage) and the original distribution (preserving representativeness), denoted as
\begin{equation}
\label{eq:q_ours_mix}
q = \alpha u + (1-\alpha)p,
\qquad \alpha \triangleq \frac{K}{Q},
\end{equation}
where $u\in\Delta^{K-1}$ denotes the uniform distribution over the $K$ observed equivalence classes, and
$u_k \triangleq 1/K$, and we focus only on the common regime $Q\ge K$ where full class coverage is feasible. Table~\ref{tab:ecs_notation} summarizes the notations.

\begin{table}[t]
\caption{Notations.}
\centering
\small
\begin{tabular}{ll}
\toprule
Symbol & Meaning \\
\midrule
$\tau$ & A fixed tool \\
$\mathcal{E}_\tau$ & Experience pool for tool $\tau$ \\
$N_\tau$ & Number of experiences, $N_\tau=|\mathcal{E}_\tau|$ \\
$d$ & Score dimension, $\kappa(E)\in\{0,1\}^d$ \\
$\mathcal{C}_k$ & $k$-th equivalence class \\
$K$ & Number of observed equivalence classes \\
$Q$ & Selection quota \\
$p$ & Class distribution of experience pool, $p_k=|\mathcal{C}_k|/N_\tau$ \\
$q$ & Class distribution of selected experiences\\
$u$ & Uniform distribution over $K$ classes, $u_k=1/K$ \\
$\alpha$ & Mixture weight, $\alpha=K/Q$ \\
\bottomrule
\end{tabular}
% \vspace{1mm}
\label{tab:ecs_notation}
\end{table}

\subsection{Metrics and Bounds}
The equivalence-class-based selection method aims to induce a distribution $q$ that (1) remains representative of the
original experience pool while (2) covering as many distinct invocation patterns as possible given quota $Q$.
We quantify these two objectives using (i) distribution alignment and (ii) diversity, and derive lower bounds to show the effectiveness of ours.

\subsubsection{Distribution Alignment}
Distribution alignment can be measured by the negative KL divergence~\cite{kl}
\begin{equation}
\text{Align}(q,p)\triangleq -D_{\mathrm{KL}}(q\|p)
= -\sum_{k=1}^{K} q_k \log\frac{q_k}{p_k},
\end{equation}
where larger values indicate better alignment. 

Under Eq.~\eqref{eq:q_ours_mix}, we lower bound $\text{Align}(q,p)$ by upper bounding $D_{\mathrm{KL}}(q\|p)$.
Using the convexity of $D_{\mathrm{KL}}(\cdot\|p)$ in its first argument, we have
\begin{align}
D_{\mathrm{KL}}(q\|p)
&= D_{\mathrm{KL}}\!\left(\alpha u + (1-\alpha)p \,\middle\|\, p\right) \nonumber\\
&\le \alpha D_{\mathrm{KL}}(u\|p) + (1-\alpha)D_{\mathrm{KL}}(p\|p) \nonumber\\
&= \alpha D_{\mathrm{KL}}(u\|p). \label{eq:kl_upper}
\end{align}
Equivalently,
\begin{equation}
\label{eq:align_lower}
\text{Align}(q,p) \ge -\alpha D_{\mathrm{KL}}(u\|p).
\end{equation}

As the quota $Q$ increases, $\alpha=K/Q$ decreases, the bound in Eq.~\eqref{eq:kl_upper} tightens, and $q$ becomes progressively closer to the original distribution $p$. 
In addition, our fine-grained adjustment reallocates any leftover quota to other classes that still have available instances. 
This avoids wasting quota due to over-allocation to small classes and helps keep the induced distribution $q$ closer to the original distribution $p$, which further improves the $\text{Align}(q,p)$. The distribution alignment $\text{Align}(q,p)$ achieves its upper bound of $0$ when the selected distribution $q$ exactly matches the original distribution $p$. 
This design preserves representativeness of the selected experiences by allocating more quota to larger equivalence classes, which correspond to more frequent invocation patterns in the original experience pool of tool $\tau$.

% 因为占比越大的等价类代表了越经常出现的调用范式，这使得选择出的经验具有代表性。
% 我们算法中的Fine-grained adjustment将多余的配额二次发送，消除了随机采样因为类别数目不够带来的空白。使得DISTRIBUTION ALIGNMENT尽可能靠近这个上限，并在q=p的时候达到最大。

\subsubsection{Diversity}
Diversity can be measured using Shannon entropy~\cite{entropy}
\begin{equation}
\text{Div}(q)\triangleq H(q)
= -\sum_{k=1}^{K} q_k \log q_k,
\end{equation}
where larger values correspond to more uniform coverage over invocation patterns. 

Under Eq.~\eqref{eq:q_ours_mix}, we can guarantee a non-trivial level of diversity by lower bounding $H(q)$.
Using the concavity of entropy,
\begin{align}
H(q)
&= H\!\left(\alpha u + (1-\alpha)p\right) \nonumber\\
&\ge \alpha H(u) + (1-\alpha)H(p) \nonumber\\
&= \alpha \log K + (1-\alpha)H(p). \label{eq:entropy_lower}
\end{align}
Therefore, our method guarantees $\text{Div}(q)$ is at least a convex combination of maximal diversity $\log K$ and the original diversity $H(p)$, controlled by $\alpha=K/Q$, and $\text{Div}(q)$ approaches its maximum value $\log K$ when the selection quota $Q$ is close to the number of equivalence classes $K$. This ensures that the selected experiences maintain broad coverage over invocation patterns even under a limited quota, which is particularly beneficial when original distribution $q$ is long-tailed. In such cases, it mitigates the dominance of a few high-frequency patterns and preserves coverage of rare yet informative ones.

\subsection{Discussion}
% 基于等价类的划分是一种扩展性很强的方法。其能够很自然地利用LLM评估得到的多维向量，将不同方面的质量综合起来，整理出不同的类别范式以供使用。继承自这种划分，基于等价类的选择能够灵活地平衡对常见范式的关注程度和对不同范式的覆盖性。该方法可以扩展至广阔的应用场景中，从而在不改变下游目标的情况下更显式地控制所选子集的代表性和多样性。比如样本选择，和损失定义。
% Equivalence class partitioning is a highly scalable approach. By aggregating quality signals across multiple dimensions, it naturally organizes the results of LLM-as-a-Judge into distinct pattern categories for downstream use. Building on this partitioning, an equivalence-class-based selection method can flexibly balance prioritizing common patterns with covering diverse ones. This method can be applied to a wide range of tasks that require explicit control over the representativeness and diversity of a selected subset, such as sample selection and loss-function design.
As a robust and inherently scalable architecture, equivalence class partitioning redefines how quality signals are aggregated. It systematically maps the outputs of LLM-as-a-Judge into a rigorous taxonomy of pattern categories. This foundational partitioning empowers a superior selection mechanism that masters the critical equilibrium between representativeness and diversity. Beyond mere subset selection, this framework serves as a universal engine for any task requiring precise distributional control, ranging from strategic data mixture to the fundamental optimization of loss functions. 
% 很自然的于工具调用的错误归因联系起来。但该形式本质上是通用的：只要能够通过某个评分函数 $\kappa(\cdot)$ 将样本映射到离散模式（等价类），并且存在固定预算 $Q$（例如上下文长度、存储容量或标注成本），就可以用相同思路进行子集构建。进一步地，在许多下游方法中，训练/优化往往在样本层面定义损失或奖励（loss/reward）并据此进行采样或加权；此时也可以引入本文的等价类视角，将原本“基于样本”的策略扩展为“先在类级别选择/分配，再映射回样本”的实现方式，从而在不改变下游目标的情况下更显式地控制代表性与覆盖度。% 基于将等价类的划分方法能够自然的将LLM评估得到的二元分数与工具调用的模式对应起来，从而支持基于等价类的

\section{Baseline Details}\label{app:baselines}
\paragraph{Few-shot Learning.} A prompting method~\cite{fewshot} that provides a small number of tool-use examples in the prompt to let the model learn from them. To ensure a fair comparison, we used the same invocations in the train set as in {\ours} as examples, and the same truncation for each tool according to the context length.
\paragraph{DRAFT.} Dynamically Refining
tool documentation through the Analysis of Feedback and Trials (DRAFT)~\cite{Draft} is a framework designed to bridge the comprehension gap between LLMs and external tools through an iterative and self-improving methodology. Unlike our approach, DRAFT decouples the process from the agent’s inference procedure and improves tool documentation quality through three steps: experience gathering, learning from experience, and documentation rewriting. Since it requires the tool to be actually executed in order to function, on the tools that cannot be actually executed in the MetaTool dataset and parts of the BFCL-v3 dataset, we use the tool invocation information we extracted in {\ours} instead of the real execution. We use Qwen3-max as the agent model, consistent with our setup.
\paragraph{Mem0.} Mem0~\cite{chhikara2025mem0} is a memory layer designed to provide agents with a persistent knowledge base. In our experiments, we specifically utilize its \texttt{procedural\_memory} module (invoked via the \texttt{add} method with \texttt{memory\_type="procedural\_memory"}). As its native design cannot reliably capture tool-level signals (it mainly stores user preferences or coarse-grained agent behaviors), we explicitly provide the agent–tool interactions from the training set. When retrieve the memory during inference, we apply the same truncation for each tool according to the context length. We use Qwen3-max as the agent model, consistent with our setup.
% By using Mem0 as a baseline, we evaluate the performance of a standard procedural memory storage and retrieval system against \ours, which introduces more complex distillation and utility-based refinement mechanisms.
\label{tried}

\section{Dataset and Metric}\label{app:dataset}
\paragraph{MetaTool} MetaTool~\citep{metatool} is a benchmark designed to evaluate whether LLMs have tool usage awareness and can correctly choose tools. 
We use four sub-task settings that test tool selection to construct a dataset centered on the 100 most frequently used tools, with a maximum of 10 tasks per sub-task involving the same tool, forming total 1669 tasks. Specifically, $\textit{Similar}$ sub-task setting denotes tool selection tasks with similar tool candidates (500 tasks); $\textit{Scenario}$ sub-task setting denotes tool selection tasks in specific scenarios (530 tasks); $\textit{Reliability}$ sub-task setting denotes tool selection tasks with possible reliability issues, where the ground truth tool is excluded from the candidate list (500 tasks); and $\textit{Multi-tool}$ sub-task setting denotes multi-tool selection tasks (139 tasks). 
For evaluation, we follow the benchmark’s matching method to assess whether the set of tools selected by the agent exactly matches the ground truth answers.

\vspace{-1mm}
\paragraph{API-Bank} API-Bank~\citep{apibank} is a benchmark designed to evaluate the end-to-end performance of tool-augmented LLMs, specifically in planning, retrieving, and invoking tools within dialogue systems. 
It simulates a comprehensive tool-use environment containing 53 diverse tools and includes 399 tasks in its Level-1 subset, which specifically targets the correctness of tool invocations. 
For evaluation, we follow the dataset’s designed matching method, focusing on precision by checking whether the predicted tool name and parameter values exactly match the ground truth annotations.
% In our experiments, we adopt the standard evaluation protocol of API-Bank Lv1 to measure the agent's success in correctly formulating API invocations based on user instructions.

\vspace{-1mm}
\paragraph{BFCL-V3} The Berkeley Function Calling Leaderboard V3 (BFCL-V3)~\citep{patilberkeley} is a comprehensive benchmark assessing tool invocation capabilities across diverse programming languages and interaction patterns. 
It comprises varied categories, ranging from simple function invocations to complex scenarios involving multiple or parallel invocations. Because some of these tools have only been invoked one time, we aggregate the tasks according to the use of the tools in the ground truth. We then filter the data to keep only tasks in which every tool used also appears in at least two other related tasks, and we randomly sample up to 50 tasks per category, excluding the ``Long Context'' category to avoid confounding effects from different context-length limits. The remaining 461 tasks constitute the test set. To build the training set, we gather all corresponding related tasks and, for each tool, randomly sample 10 tasks when it is associated with more than 10, resulting in 1,542 training tasks. In our analysis, we divide the tasks into four sub-tasks according to (1) the number of inference turns required and (2) whether the task involves problematic settings, such as testing tool-use relevance (irrelevance), missing tools, or missing required information for tool invocation. Specifically, $\textit{Normal}\mathrm{}_{\mathrm{single}}$ sub-task setting denotes single-turn baseline tasks under normal settings (255 tasks); $\textit{Normal}\mathrm{}_{\mathrm{multi}}$ sub-task setting denotes multi-turn baseline tasks under normal settings (50 tasks); $\textit{Problem}\mathrm{}_{\mathrm{single}}$ sub-task setting denotes single-turn tasks with problematic settings (56 tasks); and $\textit{Problem}\mathrm{}_{\mathrm{multi}}$ sub-task setting denotes multi-turn tasks with problematic settings (100 tasks). 
For evaluation, the benchmark employs a rigorous dual-evaluation framework: Abstract Syntax Tree matching to verify syntactic correctness and executable testing to ensure the functional outcome matches the ground truth. A task is considered successful only if all the rounds and steps of inference are judged to have passed by both.
\section{Implementation details}\label{app:tried}
\subsection{Settings about Experience}
In the main experiments, we use our designed generic tool invocation evaluation to collect experience with the 10-dimensional vector $scores$. For the MetaTool dataset, the answer has only the tool name and no real execution of the tool. Thus, we include a simple prompt ``This is a special tool selection task, you only need to evaluate the Use Quality part, the Input Quality and the Response Quality should be assumed to be perfect and get all the points by default.'' that guides the agent to focus only on tool selection. For tools in the dataset that can be executed, we use their actual execution results and returned outputs as the tool response field in metadata $m$. This returned content may be vague or even empty, which constitutes one of the challenges that {\ours} is designed to handle. For tools that cannot be executed, we instead use the ground truth as the tool response. When validating the effectiveness of our equivalence-class-based selection method, since the experiences we collected all have $d=10$, we compute the expected probability of 1 for a single dimension, and use a Bernoulli distribution~\cite{Bernoulli} to simulate cases where $d$ takes other values.

\subsection{Hyperparameter Quota $Q$}
Given a tool $\tau$, we aim to select a safe maximum quota under a fixed context-length budget $B$ (e.g., the maximum tokens allowed for ${\text{LLM}}_{\text{summarizer}}$ beyond the prompt). Let $\mathcal{E}_\tau=\{E_i\}_{i=1}^{N_\tau}$ be the experience pool for $\tau$. We define the length $\ell(E_i)$ of an experience as the token length of its serialized form used at inference time.
We first sort experiences by length in descending order. Let $\pi$ be a permutation such that
\begin{equation}
\small
\label{eq:sort_desc}
\ell\!\left(E_{\pi(1)}\right)\ge \ell\!\left(E_{\pi(2)}\right)\ge \cdots \ge \ell\!\left(E_{\pi(N_\tau)}\right).
\end{equation}
We then compute the target selection quota $Q_\tau$ via a greedy procedure that iterates from the longest experience to the shortest and stops once the budget would be exceeded, as denoted:
\begin{equation}
\small
\label{eq:quota_greedy}
Q_\tau \triangleq \max\left\{q\in\{0,\ldots,N_\tau\} \;\middle|\; \sum_{j=1}^{q}\ell\!\left(E_{\pi(j)}\right)\le B_\tau \right\}.
\end{equation}

As a result, $Q_\tau$ is the safe maximum number of experiences that can be accommodated within budget $B$, which is used for subsequent experience distillation and experience reuse phases.

\subsection{Use of Guidance} 
After obtaining guidance from {\ours}, we use it in different strategies depending on its category. For dynamic guidance, we append it to the end of the tool description using the following template: ``This is optional guidance on how to better use the tool. You may refer to it selectively: \{dynamic guidance\}.'' For stable guidance, we store it in a file and load it alongside the tool schema. It is then inserted into the schema using the prompt: ``This is guidance on using the tool learned from previous experience. You must follow it: \{stable guidance\}.''

\subsection{Things We Tried}

% 在\{ours}的设计和效果验证时，我们尝试了很多其它方法，这些方法可能没有起作用或不符合我们的设定。这是我们能记住的东西。\textbf{guidance添加位置}。我们尝试将获取到的guidance按工具列表顺序添加到角色为system的提示词最后，这略微降低了我们的效果，我们推测是因为这使得模型增加了一个“从工具列表中去system prompt找到对应工具的guidance”的过程。\textbf{将groundtruth作为采样得到的正向经验加入reme}.这使得我们的方法的效果略微有一点提升（不到1%），说明{\ours}已经可以在没有groundtruth的时候很好的总结工具能力和最佳实践。\textbf{使用更严格的tool response}。我们将数据集evalution的结果加入工具响应，并提示模型“你应该这样完成这个任务...”，这使得我们的方法有很大波动，经过人工排查发现数据集本身质量问题会提供噪声。\textbf{尝试了更多epoch.}我们在实验中的数据集上测试了使用我们的方法进行多epoch推理，每一轮会使得\{ours}得到同一个任务上不同的经验。我们发现{\ours}在第二epoch后几乎不会再有提升，说明reme能在很快时间内收敛，而无法做对的任务则反应了模型和我们方法的上限。\textbf{{使用我们在经验获取阶段获取的经验存入Mem0。}这使得Mem0有了较大进步，能够和我们的方法相比较，但原理上来看这和我们的方法区别不大且并非Mem0的官方实现。\textbf{和基于微调的方法比较。}我们还将我们的方法和两个能找到的基于微调的方法进行了比较，分别是Toolllama和huggingface上发布的qwen3-8b模型，前者因为接口的落后无法与当下任务匹配而取得了很低的分数（几乎是baseline方法的一半），后者尽管相比基础模型有了一定改进（大约3%），但因起没有披露详细的训练方法而被移除。
During the design and evaluation of {\ours}, we explored many alternative variants in the early stage. Some did not work or did not align with our experimental setting, and here are something we can remember: 
\textbf{Guidance insertion position.}
We append the generated guidance to the end of the system prompt following the order of the tool list. This slightly degrades performance. We conjecture that it introduces an extra step for the model to look up the corresponding guidance for each tool from the system prompt based on the tool list. 
\textbf{Adding ground truth as positive trajectories into {\ours}.}
Acquiring experiences from ground truth yields a small improvement (less than 1\%). This suggests that {\ours} can already summarize tool's capability boundaries and best practices well without ground truth or fine-grained feedback.
% {采样标准差} 我们统计了主实验中的三次采样的标准差，尽管在部分设定上有轻微出入，我们的方法在大多数设定上相比其它所有方法都取得了最小的标准差，从另外一个角度验证了{\ours}对agent调用工具鲁棒性的增强。
% \textbf{Using stricter tool responses.}
% We augmented tool responses with dataset evaluation results. We also instructed the model with directives such as ``you should complete this task in this way \ldots''. This led to large performance fluctuations. Manual inspection indicates that quality issues in the dataset can inject substantial noise. 
\textbf{Compare standard deviation.}
We compute the standard deviation over the three runs in our main experiment. Our method achieves the smallest standard deviation compared to all other baselines in most settings, providing additional evidence that {\ours} improves the robustness of the agent’s tool use. 
\textbf{More epochs.}
We test more epoch inference on our experimental datasets. Each epoch allows {\ours} to acquire experiences from different epoch for the same task. We observe almost no further improvement after the second epoch. This indicates that {\ours} converges quickly, and tasks that remain unsolved reflect the upper bound of the backbone agent and our method. 
% \textbf{Storing experiences acquired in the experience acquisition stage into Mem0.}
% This modification substantially improved Mem0, making it comparable to our method. However, we do not report these results because they do not align with mem0’s design and, conceptually, they amount to an alternative implementation of our approach. 
\textbf{Comparison with fine-tuning-based methods.}
We also compare our method with two available fine-tuning-based baselines. They are ToolLLM~\cite{11} and a Qwen3-8B-sft~\footnote{\url{https://huggingface.co/1nstaller/qwen3_8b_sft_tool_use}} model released on HuggingFace. ToolLLM achieves very low scores, nearly half of the ``No Method'' baseline, perhaps because its backbone LLM is outdated and does not match current tasks. The Qwen3-8B-sft shows some improvement over the base model (about 3\%), but is removed because its training details are not disclosed. \textbf{Reliability of agent judgments.}
We manually inspect the guidance generation process for a subset of tools. Although the ${\text{LLM}}_{\text{evaluator}}$ can occasionally misjudge certain aspects and assign inaccurate $scores$ during the evaluation, most of the acquired experiences (over 95\%) are reliable and informative. Moreover, imperfect experiences are likely to be corrected during experience distillation, since ${\text{LLM}}_{\text{summarizer}}$ can synthesize them from a higher-level perspective. Overall, 98.7\% of the resulting guidance are deemed reliable by human reviewers. Guidance considered unreliable can also be partially removed by the experience filtering step in subsequent epochs.  
\textbf{More challenging tasks.}
We also verify the effectiveness of our method on other benchmarks. Under the ``Oracle'' setting of MCPVerse~\cite{lei2025mcpverse} using Qwen3-32B as the backbone model, our method improves the average accuracy from 23.00\% to 26.10\%. However, for smaller models, the gains are substantially constrained by agent capability and context limitations. To more clearly show our method’s gains in tool use, we do not choose these challenging tasks as our main experimental benchmarks. \textbf{Alternative design variants.}
We also explore several alternative variants across different phases of {\ours} to better understand the impact of each design choice. Each row replaces exactly one component of the original design without combining modifications. The results are summarized in Table~\ref{tab:variants}.

% agent可信度。我们对一部分工具的guidance生成过程进行了人工审查。尽管一些时候在经验获取阶段的agent_{evaluator}评分不够准确，可能在某个方面判断失误，但是大部分获取出的经验（95%以上）都是可靠且言之有物的。对于那些有瑕疵的经验，在经验蒸馏阶段也有很大概率被纠正，因为agent_{summarizer}能够从更高层的角度进行判断。最终，得到的guidance被评审人员认为是可靠的占比达到了98%。那些被认为是不可靠的guidance，也在后续阶段的experience filtering 步骤中有过删除。{更有挑战性的任务。}我们在mcp相benchmark上也验证了我们方法的有效性，在mcpverse的Oracle设定中，使用Qwen3-32B模型作为backbone，我们的方法将平均准确率从23.00%提高到26.10%体现了其有效性。然而在较小尺寸的模型上，该提升大大受到agent能力和上下文的制约，为了验证我们方法的通用性，我们没有选择这些对agent而言难度很大的任务作为实验benchmark。

\begin{table*}[t]
\centering
\caption{Ablation study of alternative variants across different phases of {\ours}. All numbers are Pass@1 on BFCL-V3 with Qwen3-8B. Each row replaces one component of the original design.}
\resizebox{0.9\linewidth}{!}{
\begin{tabular}{llcc}
\toprule
\textbf{Phase} & \textbf{Variant} & \textbf{Accuracy} & \textbf{Reduction vs Original} \\
\midrule
\multirow{1}{*}{Original}
& Full method & \textbf{67.46} & 0.00 \\
\midrule
\multirow{2}{*}{Experience Acquisition}
& Directly use full trajectories (no experiences) & 60.08 & -7.38 \\
& Directly prompt the LLM to extract experiences & 63.34 & -4.12 \\
\midrule
\multirow{4}{*}{Experience Distillation}
& Use only the most recent 1 experience & 62.26 & -5.20 \\
& Keep only the most recent 3 experiences & 64.64 & -2.82 \\
& Remove the experience filter step & 65.94 & -1.52 \\
& Remove the experience summary step & 63.56 & -3.90 \\
\midrule
\multirow{2}{*}{Experience Reuse}
& Only dynamic strategy & 66.37 & -1.09 \\
& Only stable strategy & 63.56 & -3.90 \\
\bottomrule
\end{tabular}
}
\label{tab:variants}
\vspace{-1mm}
\end{table*}

% \newpage
\section{Case Analysis}\label{case}

\subsection{{\ours} Improves the Agent's Tool-Use Ability from Multiple Aspects}
We summarize representative cases of tool-use improvements brought by our method into six aspects. For each aspect, we provide an illustrative example with a brief explanation.

\paragraph{(1) Improve Tool-Use Awareness.}

When the user implicitly asks about something related to a historical period, the agent tends to answer directly from its internal knowledge in a conversational manner instead of invoking the \texttt{timeport} tool. {\ours} learns from past experience that the tool should be invoked in such cases.
\begin{figure}[!h]
    \centering
    \includegraphics[width=0.98\linewidth]{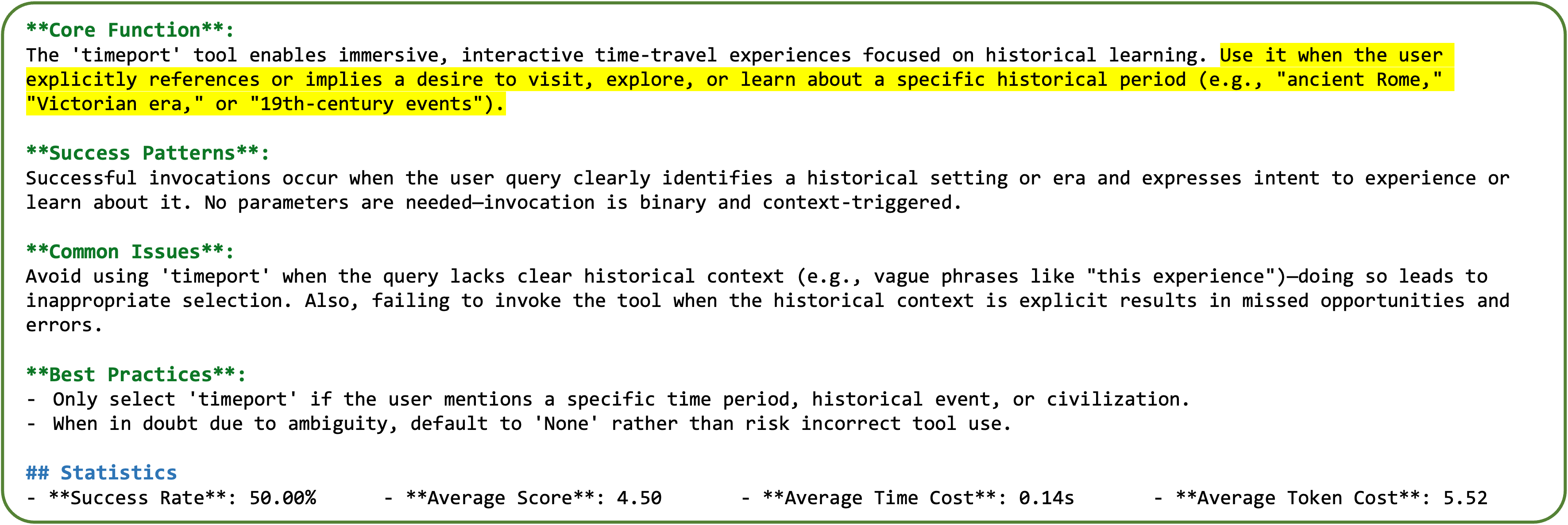}
    \caption{Successful case of improved tool use: improve tool-use awareness.}
    \vspace{-2mm}
\end{figure}

\paragraph{(2) Distinguish Similar Tools.}
The agent uses \texttt{RestaurantBookingTool} to make restaurant reservations. When the user's request is ambiguous, the agent often confuses it with the semantically similar \texttt{local} tool, since \texttt{local} can be used to search for restaurants. {\ours} learns the capability boundary from past experience, i.e., it should invoke the booking tool only when the user explicitly asks to reserve a table.
\begin{figure}[!h]
    \centering
    \includegraphics[width=0.98\linewidth]{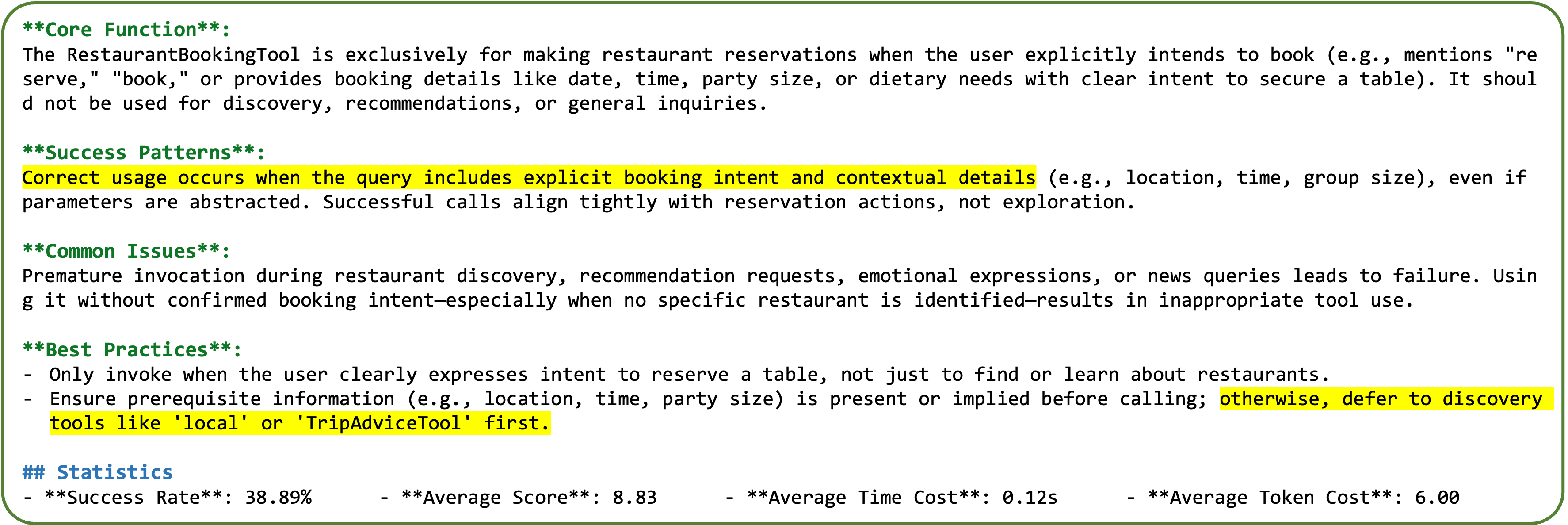}
    \caption{Successful case of improved tool use: distinguish similar tools.}
    \vspace{-2mm}
\end{figure}

\paragraph{(3) Invoke Tools Correctly.}
The agent uses the \texttt{inventory\_management} tool to complete inventory management tasks. {\ours} learns from past experience the constraints on the allowed parameter content and required formats for this tool.
\begin{figure}[!h]
    \centering
    \includegraphics[width=0.98\linewidth]{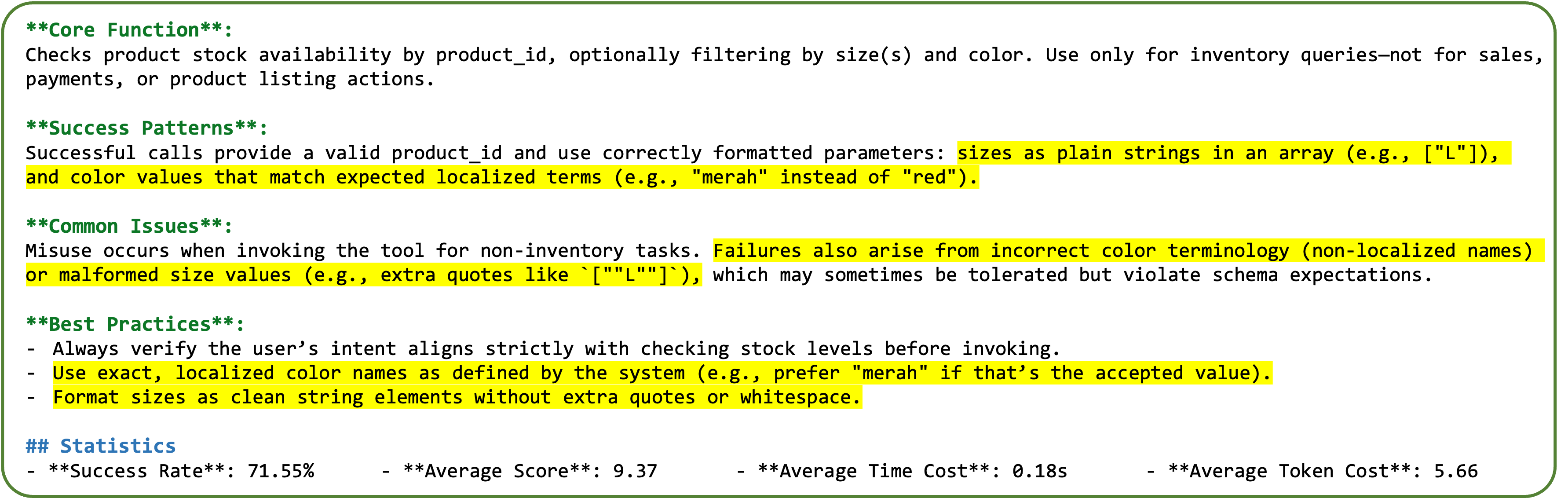}
    \caption{Successful case of improved tool use: invoke tools correctly.}
    \vspace{-2mm}
\end{figure}

\paragraph{(4) Fix Tool-Use Errors.}
The agent uses \texttt{EmergencyKnowledge} to retrieve possible diseases, but in the first invocation is failed as the agent provides an overly descriptive input. After {\ours} recognizes this issue, it fixes the input formulation in subsequent invocations.
\begin{figure}[!h]
    \centering
    \includegraphics[width=0.98\linewidth]{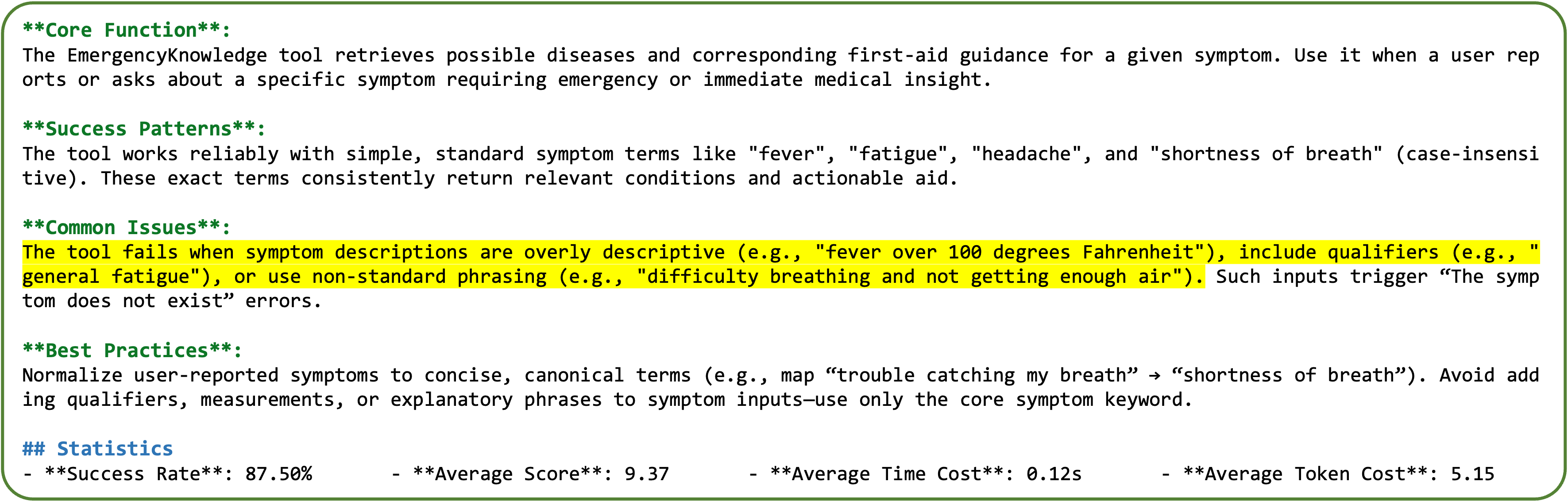}
    \caption{Successful case of improved tool use: fix tool-use errors.}
    \vspace{-2mm}
\end{figure}

\paragraph{(5) Identify Tool Dependencies.}
The agent uses the file-system tool \texttt{wc} to count the number of lines in a file. {\ours} learns from past experience that the error ``file does not exist'' occurs frequently. The guidance encourages the agent to run \texttt{ls} or \texttt{find} beforehand, which can effectively avoid such failures. 
\begin{figure}[!h]
    \centering
    \includegraphics[width=0.98\linewidth]{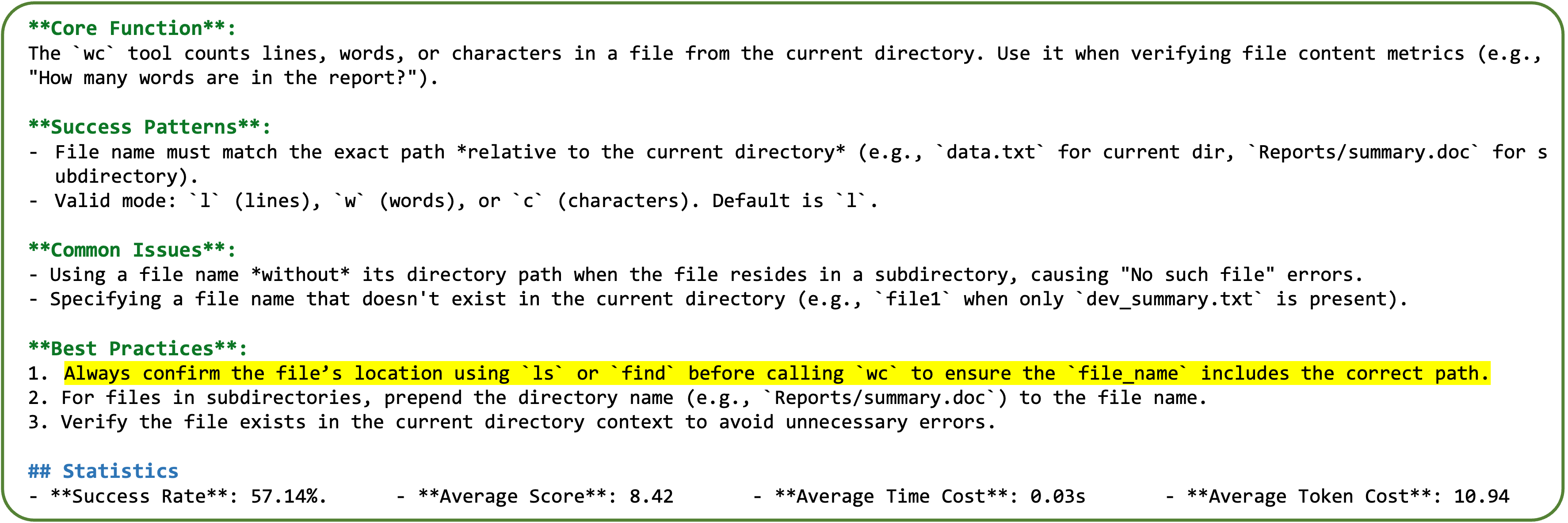}
    \caption{Successful case of improved tool use: identify tool dependencies.}
    \vspace{-4mm}
\end{figure}

\paragraph{(6) Understand Tool Reliability.}
From prior experiences where the agent uses \texttt{QueryRegistration} to check patient appointment registrations, {\ours} observes frequent failures caused by tool availability issues and identifies them as tool-side problems. It therefore reminds the agent to check tool availability before invocation or to seek alternative tools. 
% In a later query, the agent completes the task using an alternative tool.
\begin{figure}[!h]
    \centering
    \includegraphics[width=0.98\linewidth]{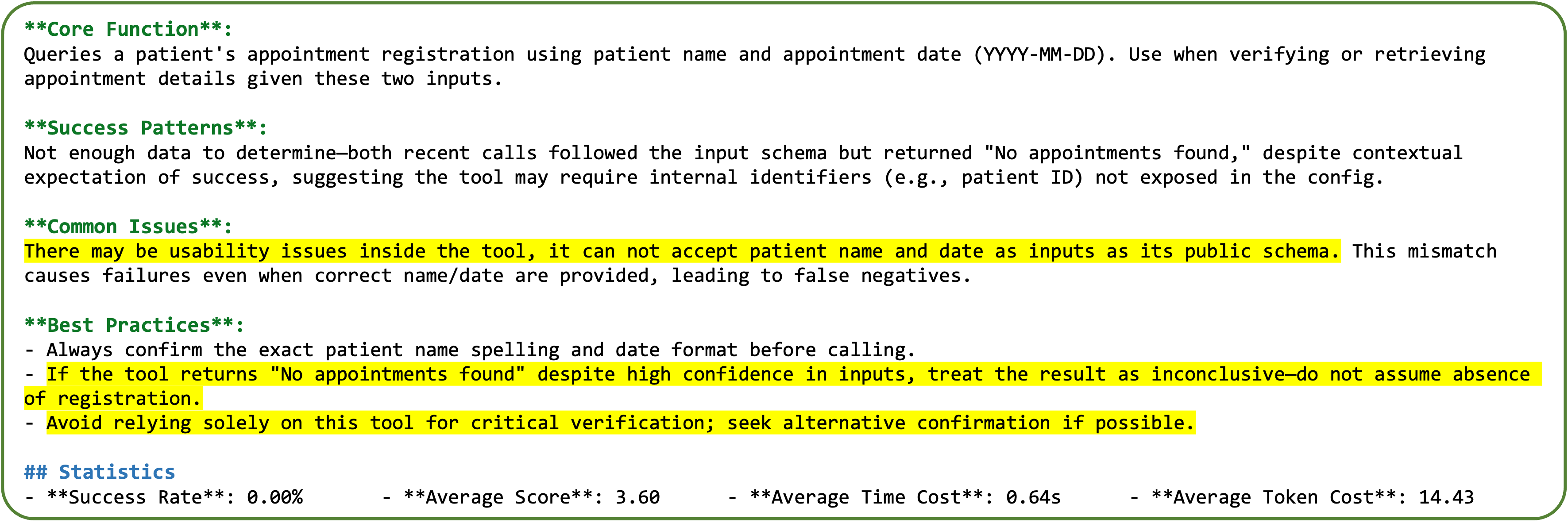}
    \caption{Successful case of improved tool use: understand tool reliability.}
    \vspace{-2mm}
\end{figure}

\subsection{Cases Where {\ours} Fails}

\paragraph{(1) Unseen Situations.}
If the agent uses an incorrect tool, or fails to use a tool when it should, {\ours} cannot collect trajectories of the correct tool invocation and therefore cannot learn from it to improve. As a result, there is no experience or guidance for that tool. For previously seen tools, {\ours} may still fail when experience is limited: some invocation patterns or error modes may not have appeared in the past trajectories, leaving the agent without relevant guidance to prevent them.
% \begin{figure}[!h]
%     \centering
%     \includegraphics[width=0.98\linewidth]{latex/figures/badcase1.png}
%     \caption{Case of failed tool use: Unseen Tools.}
% \end{figure}

\paragraph{(2) Hard-to-analyze Trajectories.}
% A user asks the agent to set an account password. After multiple rounds of tool interaction, the agent finds that no account exists and decides to register one. This logic may be reasonable for many agents, so {\ours} also fails to summarize that the user intent is merely to initialize the account; if no account exists, the agent should further communicate with the user. Such issues are stable to capture because they stem from the agent's ability to precisely infer user expectations in every step.
% agent调用edit_ticket完成用户“帮我编辑3号票据内容为出货信息。”的任务。然而智能体判断的出货信息包括所有的货物名称，数量和价格明细等信息，与用户所期望的符合模板等物流信息不同，所以即使该工具被成功调用，用户的要求仍然没有完成。该错误是由于智能体对用户意图的判断失误而不是工具调用错误而引起的，属于{\ours} 可以解决范围之外的问题。
The agent invokes \texttt{edit\_ticket} to handle the user request: ``Please edit ticket No. 3 to contain shipping information.'' However, the agent interprets ``shipping information'' as including full details such as all item names, quantities, and price breakdowns, which differs from what the user expected information formatted according to a template. As a result, even though the tool invocation is successful, the user’s request is still not satisfied. This error is caused by the agent’s misinterpretation of the user’s intent rather than a tool invocation failure, and it falls outside the scope of issues that {\ours} can address.
\begin{figure}[!h]
    \centering
    \includegraphics[width=0.98\linewidth]{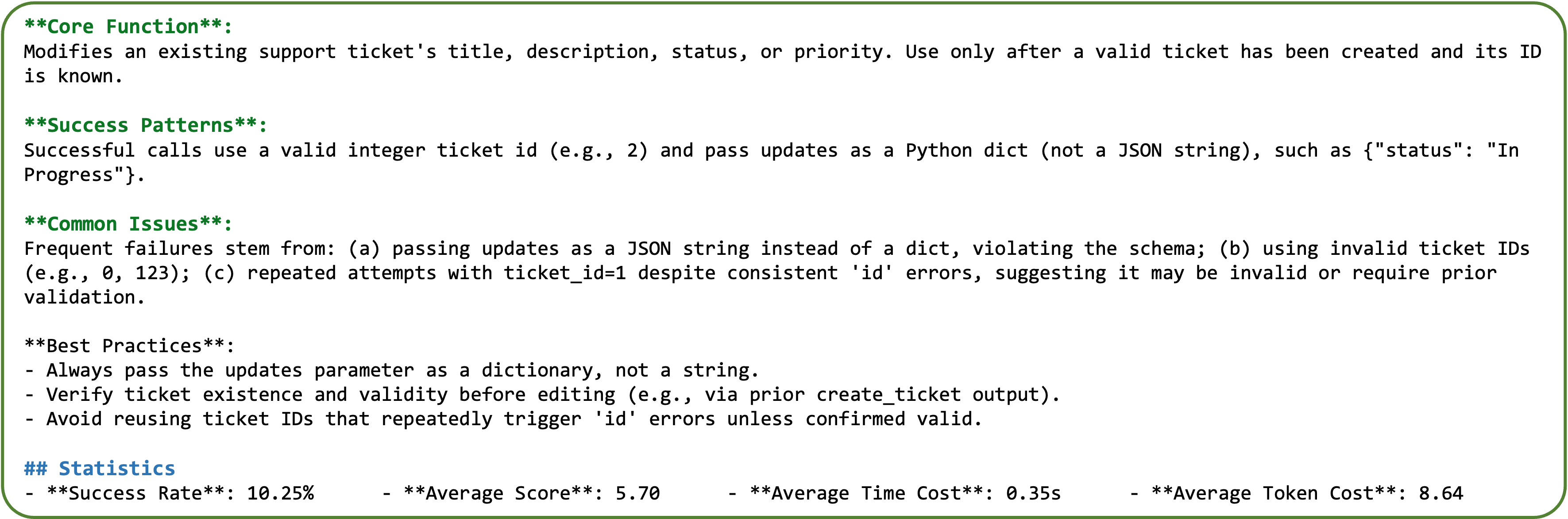}
    \caption{Case of failed tool use: hard-to-analyze trajectories.}
    \vspace{-4mm}
\end{figure}

\paragraph{(3) Incorrect Attributions.}
% The agent uses \texttt{CancelTimedSwitch} to adjust device status. Because the tool's actual behavior is inconsistent with its schema, the agent never succeeds in invoking it. However, the tool always returns parameter-related errors. During summarization, {\ours} identifies it as a tool-side problem, but in the next attempt it incorrectly attributes the failure to the agent's inappropriate behavior rather than the tool's own issue.
The agent adjusts the device state using \texttt{CancelTimedSwitch}. Because the actual behavior of the tool is not consistent with its schema (the schema specifies that the device id should be passed as the content of parameter ``device\_name''), the agent will never successfully invoke it. However, the tool always returns a ``device not present'' response because it can't find the device. {\ours} incorrectly attributes the failure to inappropriate behavior of the agent, as the agent did not pass in the correct device name, rather than a problem with the consistency of the tool itself.
\begin{figure}[!h]
    \centering
    \includegraphics[width=0.98\linewidth]{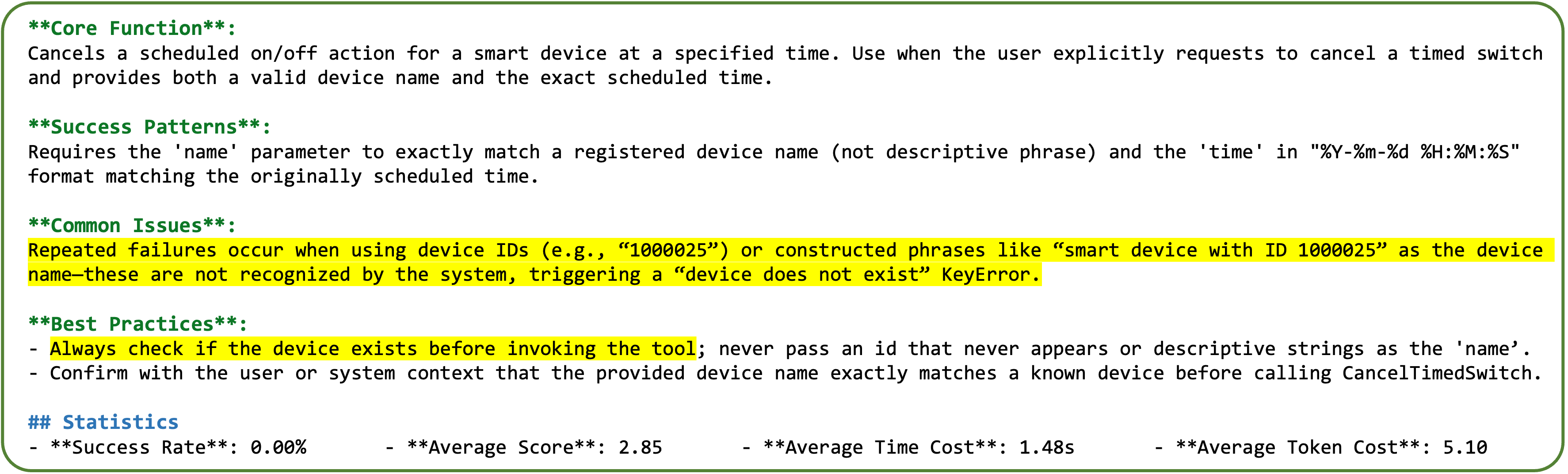}
    \caption{Case of failed tool use: incorrect attributions.}
    \label{fig:badcase}
    \vspace{-2mm}
\end{figure}

\paragraph{(4) Over-Reaction.}
When doctors’ names always included a prefix, the agent was able to use \texttt{ModifyRegistration} successfully in past experience. Consequently, {\ours} inferred a piece of guidance: always use the name with a title. However, those earlier successes occurred only because the doctor names stored in the system already contained the prefix. When a doctor is entered with a full name rather than a titled name, the agent, following the guidance, still forcibly adds the ``Dr.'' prefix, which leads to a failure.
\begin{figure}[!h]
    \centering
    \includegraphics[width=0.98\linewidth]{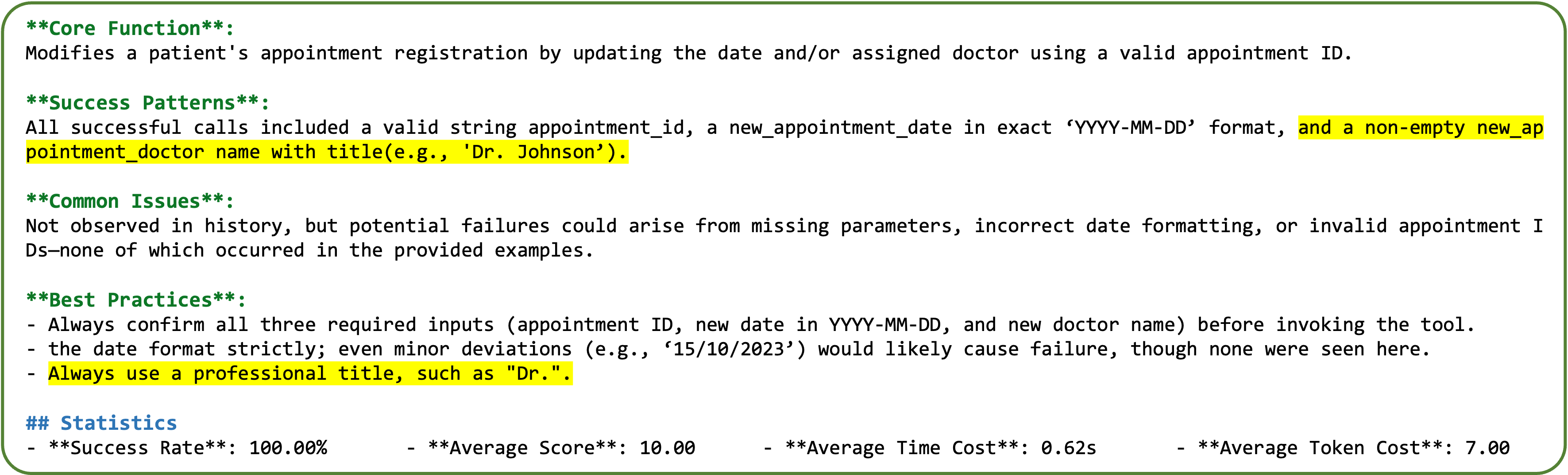}
    \caption{Case of failed tool use: over-reaction.}
    \vspace{-2mm}
\end{figure}

\subsection{Case Coverage Comparison}
% 以API-bank为例，我们选了其中一次推理结果，对{\ours}和其它方法进行了Case-coverage Comparison。图~\ref{overlap}展示了它们相比没有方法的baseline做对和做错样本的数量。在比baseline表现好的部分，我们的方法覆盖了大部分的Few-shot和DRAFT能做对的样本，但和mem0这种对agent进行增强的方法有互补优势。在比baseline Underperform的部分，明显地我们的方法相比其它方法引入了更少的错误，归功于我们Experience Distillation phase对经验进行了effectiveness check并剔除unhelpful的经验。
\begin{figure}[!h]
    \centering
    \includegraphics[width=0.7\linewidth]{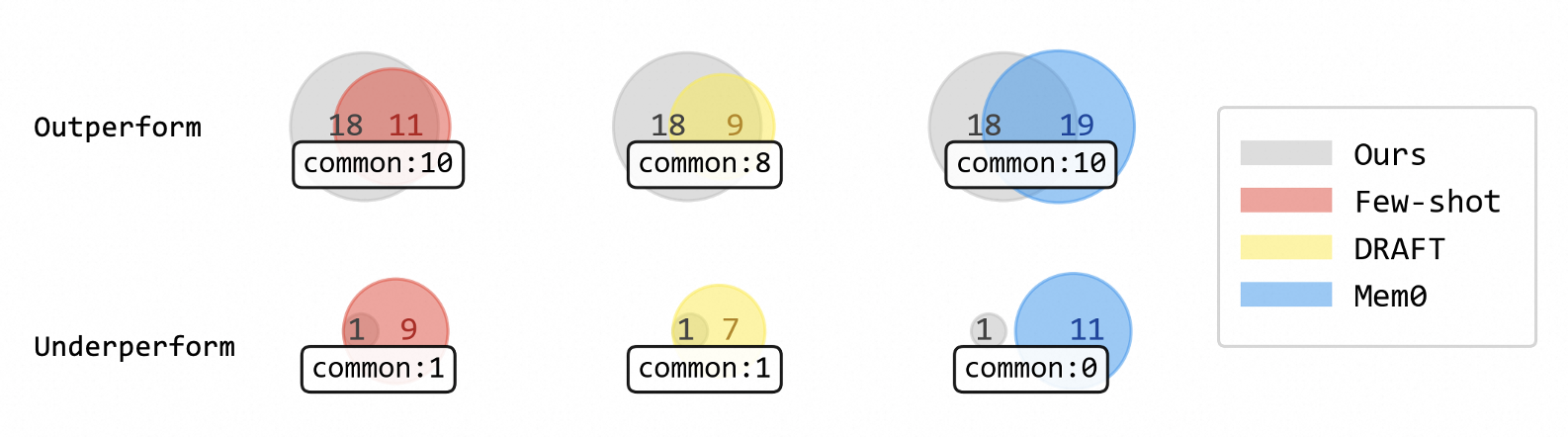}
    \caption{Case coverage comparison with baseline methods.}
    \label{overlap}
\end{figure}
Using API-Bank as the dataset, we select one run as a representative result and compare case coverage between {\ours} and other methods. Figure~\ref{overlap} shows the numbers of cases that each method gets right or wrong compared to the ``No Method'' baseline. 
In the subset where methods outperform the baseline, {\ours} covers most of the cases solved by Few-shot and DRAFT, while exhibiting complementary gains to Mem0, possibly because Mem0 enhances the agent from a process perspective rather than focusing on tool invocation. In the subset where methods underperform the baseline, {\ours} introduces substantially fewer error cases than the other methods. We attribute this robustness to the experience distillation phase, where {\ours} performs an effectiveness check and filters out unhelpful experiences.
%%%%%%%%%%%%%%%%%%%%%%%%%%%%%%%%%%%%%%%%%%%%%%%%%%%%%%%%%%%%%%%%%%%%%%%%%%%%%%%
%%%%%%%%%%%%%%%%%%%%%%%%%%%%%%%%%%%%%%%%%%%%%%%%%%%%%%%%%%%%%%%%%%%%%%%%%%%%%%%

\begin{table}[t]
\caption{Overhead of {\ours} on the three main benchmarks (Qwen3-8B, parallelism $=4$).}
\centering
\resizebox{0.6\linewidth}{!}{%
\begin{tabular}{lcccc}
\toprule
\textbf{Phase} & \textbf{Extra LLM calls} & \textbf{Tokens} & \textbf{Cost (CNY)} & \textbf{Time} \\
\midrule
Acquisition & 3,829 & $\approx 6.8\times10^{6}$ & 20.7 & 85\,min \\
Distillation & 3,335 & $\approx 7.2\times10^{6}$ & 24.9 & 74\,min \\
\midrule
Total & 7,164 & $\approx 1.4\times10^{7}$ & 45.6 & 159\,min \\
\bottomrule
\end{tabular}%
}
\label{tab:cost}
\vspace{-2mm}
\end{table}

\section{Cost Analysis}\label{app:compute}
\paragraph{Full reproduction.}
Full reproduction requires valid API keys for the commercial LLM endpoints referenced in the Experiments section. All backbone, evaluator, and summarizer inference is obtained through these APIs. A CPU-only machine is sufficient for orchestration and local preprocessing. 
In our setup, the driver process, dataset input and output, logging, and a lightweight vector index for storing experiences typically remained below approximately 16\,GB of RAM at peak usage. With reduced caching or smaller concurrent batches, 8\,GB of RAM is often sufficient. However, the risk of out-of-memory errors increases if many trajectories are materialized simultaneously. 
Wall-clock time is primarily determined by provider latency, API rate limits, and the specific models or ablations that are enabled. In our environment, running the full set of main experiments required approximately 12 days from start to completion. This duration may vary substantially across different environments and API service tiers. 
Reproducing all tables and baselines reported in the paper is expected to require more than $8\times10^{5}$ remote API or model calls in total.

\paragraph{Online overhead of {\ours}.}
In online use, {\ours} does not rely on embeddings or similarity matching, and experience-pool maintenance is almost only file storage and I/O.
On average, each experience takes about 5.7\,KB and each tool about 44\,KB, while a retrieve and an update take about 47\,ms and 170\,ms, respectively.
The extra LLM overhead mainly comes from two calls by the evaluator and the summarizer, with an average extra prompt length of 691 words.
Table~\ref{tab:cost} reports the measured overhead: each tool invocation incurs only an extra \$0.0018 and about 2.5 seconds.
In resource-constrained settings, we recommend using {\ours} in a ``generate once, reuse many times'' manner: because the guidance is synthesized from multiple past experiences, it tends to be broadly applicable, making it possible to avoid these extra costs while maintaining comparable performance.

\end{document}